%% file: thesis.tex
\documentclass[12pt,twoside]{diss}
\include{packages}

\include{commands}
\graphicspath{{./paper1/}{./paper2/}{./paper3/}{./paper4/}}

\begin{document}

%\pagenumbering{roman}
\hypertarget{title}{}
\pdfbookmark[0]{Title Page}{title}
\includepdf[pages={1-2}]{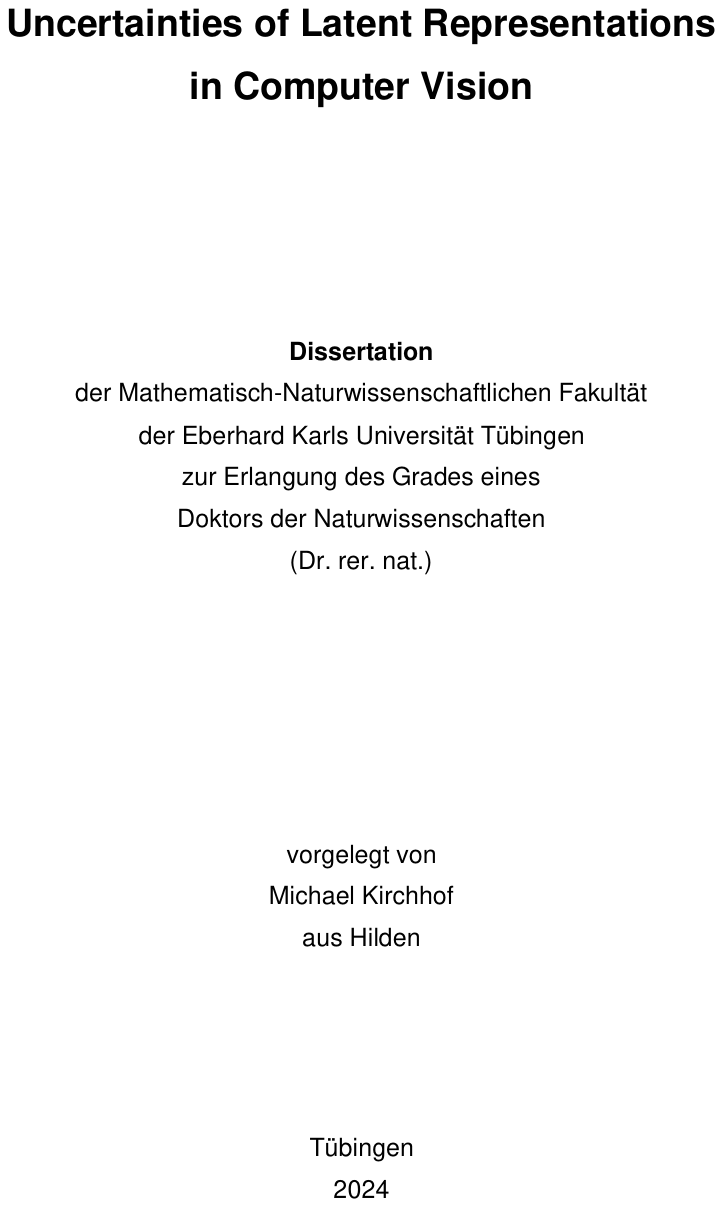}
\cleardoublepage

\hypertarget{dedication}{}
\pdfbookmark[0]{Dedication}{dedication}
\include{dedication}

\cleardoublepage

\hypertarget{abstract}{}
\pdfbookmark[0]{Abstract}{abstract}
\include{abstract}
\cleardoublepage

\hypertarget{zusammenfassung}{}
\pdfbookmark[0]{Zusammenfassung}{zusammenfassung}
\include{zusammenfassung}
\cleardoublepage

\hypertarget{acknowledgements}{}
\pdfbookmark[0]{Acknowledgements}{acknowledgements}
\include{acknowledgements}
\cleardoublepage

\dominitoc
\hypertarget{toc}{}
\setcounter{tocdepth}{1}
\pdfbookmark[0]{\contentsname}{toc}
\tableofcontents
\clearoddpage  

\include{introduction}

\clearoddpage

%\include{related_work}
%\clearoddpage

\input{paper1}
\clearoddpage

\input{paper2}
\clearoddpage

\input{paper3}
\clearoddpage

\input{paper4}
\clearoddpage

\include{conclusion}
\clearoddpage

%\addcontentsline{toc}{chapter}{\numberline{}\listalgorithmname}
%\listofalgorithms
%\clearoddpage 

\addcontentsline{toc}{chapter}{\numberline{}Abbreviations}
\setcounter{lotdepth}{2}
\listofabbrevs

\renewcommand{\arraystretch}{1.3}
\begin{tabular}{rl}
    \textbf{API} & Application Programming Interface \\
\textbf{AUROC} & Area Under the Receiver Operating Characteristic, measure \\
\textbf{Caltech 101} & Caltech 101 dataset \citep{FeiFei2004LearningGV} \\
\textbf{CARS} & Stanford Cars 196 dataset \citep{cars196} \\
\textbf{CE} & Class Entropy, uncertainty estimator \\
\textbf{CIFAR-10} & Canadian Institute For Advanced Research dataset \citep{Krizhevsky09learningmultiple} \\
\textbf{CIFAR-100} & Canadian Institute For Advanced Research dataset \citep{Krizhevsky09learningmultiple} \\
\textbf{CUB} & Caltech-UCSD Birds-200-2011 dataset \citep{cub200} \\
\textbf{DTD} & Decribable Textures dataset \citep{cimpoi14describing} \\
\textbf{ECCV} & European Conference on Computer Vision \\
\textbf{ELK} & Expected Likelihood Kernel \citep{jebara2003bhattacharyya} \\
\textbf{GPU} & Graphics Processing Unit \\
\textbf{HET-XL} & Large Heteroscedastic Classifier \citep{collier2023massively} \\
\textbf{HIB} & Hedged Instance Embeddings \citep{oh2018modeling} \\
\textbf{ICML} & International Conference on Machine Learning \\
\textbf{InfoNCE} & Info Noise Contrastive Estimation loss \citep{oord2018representation} \\
\textbf{Losspred} & Loss prediction \citep{kirchhof2023url} \\
\textbf{MCDropout} & Monte-Carlo Dropout \citep{gal2016dropout} \\
\textbf{MCInfoNCE} & Monte-Carlo InfoNCE \citep{kirchhof2023probabilistic} \\
\textbf{MLP} & Multi-layer Perceptron \\
\textbf{NeurIPS} & Neural Information Processing Systems conference \\
\textbf{nivMF} & non-isotropic von Mises-Fisher distribution \citep{kirchhof2022non} \\
\textbf{Oxford Flowers} & 102 Category Flower dataset \citep{Nilsback08} \\
\textbf{Oxford Pets} & Oxford-IIIT Pet dataset \citep{parkhi12a} \\
\textbf{PCGrad} & Projecting Conflicting Gradients \citep{yu2020gradient} \\
\textbf{ProxyNCA} & Proxy Noise Contrastive Estimation \citep{proxynca} \\
\textbf{R-AUROC} & Representation AUROC, measure \citep{kirchhof2023url} \\
\end{tabular}

\begin{tabular}{rl}
\textbf{ResNet} & Residual Neural Network \citep{resnet} \\
\textbf{RQ} & Research Question \\
\textbf{SNGP} & Spectral-normalized Neural Gaussian Process \citep{liu2020simple} \\
\textbf{SOP} & Stanford Online Products dataset \citep{song2016deep} \\
\textbf{StopGrad} & Gradient Stopping module \\
\textbf{SUN} & Scene Recognition Benchmark Database \citep{Xiao2010} \\
\textbf{SVHN} & Street View House Numbers dataset \citep{Netzer2011} \\
\textbf{Treeversity} & Treeversity dataset, single label \citep{schmarje2022one} \\
\textbf{URL} & Uncertainty-aware Representation Learning benchmark \citep{kirchhof2023url} \\
\textbf{ViT} & Vision Transformer \citep{vit} \\
\textbf{vMF} & von Mises Fisher distribution \citep{fisher1953dispersion} \\
\textbf{VTAB} & Visual Task Adaptation Benchmark, dataset collection \citep{zhai2020largescale} \\
\end{tabular}
\clearoddpage

\addcontentsline{toc}{chapter}{\numberline{}\listfigurename}
\setcounter{lofdepth}{2}
\listoffigures
\clearoddpage 

%\addcontentsline{toc}{chapter}{\numberline{}\listtablename}
%\setcounter{lotdepth}{2}
%\listoftables
%\clearoddpage

\begingroup\small
\addcontentsline{toc}{chapter}{\numberline{}\bibname}
\bibliography{bibliography}
\endgroup
\cleardoublepage

\input{appendix_arxiv}
%\clearoddpage

\end{document}

%% file: commands.tex
\makeatletter
\DeclareRobustCommand\onedot{\futurelet\@let@token\@onedot}
\def\@onedot{\ifx\@let@token.\else.\null\fi\xspace}
\makeatother

\makeatletter
\renewcommand{\paragraph}{%
	\@startsection{paragraph}{4}%
	{\z@}{0.75ex \@plus 0.5ex \@minus .25ex}{-1em}%
	{\normalfont\normalsize\bfseries}%
}
\makeatother

%% file: dedication.tex
\thispagestyle{empty}
\begin{center}
	\vspace*{\fill}
	{\hspace*{\fill} \large \textit{To whom it may concern.}}
	\vspace*{\fill}
\end{center}

%% file: abstract.tex
\thispagestyle{plain}
\chapter*{Abstract} 

Uncertainty quantification is a key pillar of trustworthy machine learning. It enables safe reactions under unsafe inputs, like predicting only when the machine learning model detects sufficient evidence, discarding anomalous data, or emitting warnings when an error is likely to be inbound. This is particularly crucial in safety-critical areas like medical image classification or self-driving cars.  Despite the plethora of proposed uncertainty quantification methods achieving increasingly higher scores on performance benchmarks, uncertainty estimates are often shied away from in practice. Many machine learning projects start from pretrained latent representations that come without uncertainty estimates. Uncertainties would need to be trained by practitioners on their own, which is notoriously difficult and resource-intense.

This thesis makes uncertainty estimates easily accessible by adding them to the latent representation vectors of pretrained computer vision models. Besides proposing approaches rooted in probability and decision theory, such as Monte-Carlo InfoNCE (MCInfoNCE) and loss prediction, we delve into both theoretical and empirical questions. We show that these unobservable uncertainties about unobservable latent representations are indeed provably correct. We also provide an uncertainty-aware representation learning (URL) benchmark to compare these unobservables against observable ground-truths. Finally, we compile our findings to pretrain lightweight representation uncertainties on large-scale computer vision models that transfer to unseen datasets in a zero-shot manner. 

Our findings do not only advance the current theoretical understanding of uncertainties over latent variables, but also facilitate the access to uncertainty quantification for future researchers inside and outside the field. As downloadable starting points, our pretrained representation uncertainties enable a range of novel practical tasks for straightforward but trustworthy machine learning.

%% file: zusammenfassung.tex
\thispagestyle{plain}
\chapter*{Zusammenfassung}

Die Quantifizierung von Unsicherheiten ist ein Grundpfeiler des vertrauenswürdigen maschinellen Lernens. Sie ermöglicht sichere Reaktionen bei unsicheren Eingaben, wie etwa Vorhersagen nur dann zu treffen wenn ein künstlich intelligentes Modell genügend Anhaltspunkte findet, anomale Daten zu filtern oder Warnungen auszugeben wenn ein Fehler wahrscheinlich ist. Dies ist besonders in sicherheitskritischen Bereichen wie der Klassifizierung medizinischer Bilder oder bei selbstfahrenden Autos wichtig. Trotz der Fülle an publizierten Methoden zur Quantifizierung von Unsicherheiten, die in numerischen Vergleichen immer bessere Ergebnisse erzielen, werden Unsicherheitsschätzungen in der Praxis oft gescheut. Viele Projekte des maschinellen Lernens starten mit vortrainierten latenten Repräsentationen, die  von sich aus keine Unsicherheitsschätzungen beinhalten. Die Unsicherheiten müssten von Anwendern selbst trainiert werden, was jedoch als kompliziert und ressourcenintensiv angesehen wird.

In dieser Doktorarbeit werden Unsicherheitsschätzer leichter zugänglich gemacht, indem sie zu den latenten Repräsentationsvektoren von vortrainierten Modellen für die Verarbeitung von Bilddaten hinzugefügt werden. Wir entwickeln Ansätze aus der Wahrscheinlichkeits- und Entscheidungstheorie, wie Monte-Carlo InfoNCE (MCInfoNCE) und die Schätzung von Vorhersagefehlern, und befassen uns sowohl mit mathematischen als auch empirischen Aspekten des Problems. Wir zeigen, dass diese unbeobachtbaren Unsicherheiten über unbeobachtbare latente Repräsentationen tatsächlich beweisbar korrekt sind. Wir stellen außerdem einen numerischen Leistungstest für Unsicherheiten über latente Repräsentationen vor (URL), um diese unbeobachtbaren Schätzungen mit beobachtbaren Vergleichswerten abzugleichen. Schließlich bündeln wir unsere Ergebnisse, um kostengünstige Unsicherheitsschätzer für die latenten Repräsentationen großer Modelle des computergestützten Sehens vorzutrainieren, die ohne weiteres Training auf neuen Datensätzen funktionieren.

Unsere Ergebnisse erweitern nicht nur das aktuelle theoretische Verständnis von Unsicherheiten über latente Variablen, sondern erleichtern auch den Zugang zur Quantifizierung von Unsicherheiten für zukünftige Forschung innerhalb und außerhalb des Feldes. Als herunterladbare Ausgangspunkte ermöglichen unsere vortrainierten Repräsentationsunsicherheiten eine Reihe neuartiger praktischer Anwendungen für unkompliziertes, aber vertrauenswürdiges maschinelles Lernen.

%% file: acknowledgements.tex
\thispagestyle{plain}
\chapter*{Acknowledgements}

While the reader may rightfully consider this thesis a document, to me it resembles a chapter of my life. Though words alone can barely express my gratitude, I want to dedicate this page to those who accompanied me through the past three years.

First, I want to thank my supervisors Enkelejda Kasneci and Seong Joon Oh. You gave me the freedom to explore my interests and guided them into well-rounded projects, with an eye for both what's theoretically intriguing and what's practically needed. I want to thank you for your mentorship beyond science, shaping me as the person I now am. All the schools, conferences, and retreats that you let me discover gave me exciting insights into the worldwide machine learning community in these exciting times. In these terms, I would like to thank the Deutsche Forschungsgemeinschaft (DFG, German Research Foundation) for funding me and my projects under Germany’s Excellence Strategy – EXC number 2064/1 – Project number 390727645. I also want to thank my third thesis advisory committee member, Wieland Brendel. Your rich feedback and deep exchange about my papers far exceeded what I hoped for. Finally, I want to thank Philipp Hennig for volunteering his time to be the independent reviewer of this thesis.

I thank my co-authors Bálint Mucsányi, Karsten Roth, Mark Collier, Tobias Leemann, Yao Rong, Elisa Nguyen, Alexander Rubinstein, and Gjergji Kasneci. I very much enjoyed the exchange and discussions, which defined what science feels like to me. I am also grateful to Bálint Mucsányi and Patrick Köhler for allowing me to supervise their master's theses. I hope I could teach you even just a fraction of what you taught me.

As evidenced by the anecdotes throughout the thesis, Tübingen was a place of collaboration to me. I particularly want to thank the International Max Planck Research School for Intelligent Systems (IMPRS-IS) for all the community events and retreats that gave space for casual conversations with hundreds of Ph.D. students working on cutting-edge papers across the whole spectrum of machine learning. This is what I dreamed of when I applied to the University of Tübingen.

Last, but most importantly, I would like to thank those that gave me joy in the best times and strength in the toughest times of my life. Above all, these are my parents Gabi Kirchhof and Harry Kirchhof and my brother Christian Kirchhof. Your unconditional support is an unwavering anchor in my life. This also holds for the best companions I could think of for this journey: Kajetan Wojtowicz, Lisa Haxel, Rachel Rapp, Torben Rapp, Patrik Reizinger, Frederic Bootz, and Florian Grieskamp. Thank you for the countless game evenings, daring outdoor adventures, and deep talks. 

This document is not long enough to list all those that had an impact on me. Freely quoting Viktor Frankl, we radiate into our effects on others.\footnote{\textit{Man's Search for Meaning}, Viktor Frankl, 1946, Beacon Press.} Rest assured that I will forever embody your effects -- figuratively for your practiced values, and quite literally for all your delicious food!

%% file: introduction.tex
\chapter{Introduction}
\label{chap:introduction}

\section{Of Penguins and Uncertainties}
\begin{figure}[t]
    \centering
    %\captionsetup[subfigure]{oneside,margin={0.8cm,0cm}}
    \begin{subfigure}{0.47\textwidth}
    \centering
        \includegraphics[width=0.7\linewidth]{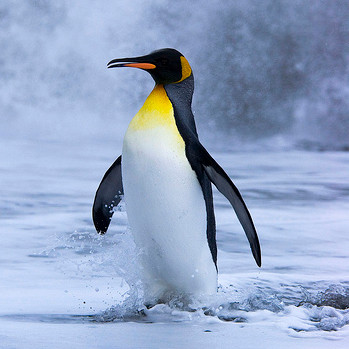}
        \caption{A clear image of a penguin.} 
        \label{fig:penguin_clear}
    \end{subfigure}%
    \hfill{\hbox{}}   % maximize separation between the subfigures
    \begin{subfigure}{0.47\textwidth}
        \centering
        \includegraphics[width=0.7\linewidth]{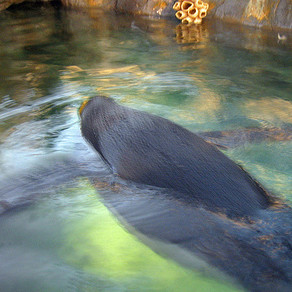}
        \caption{An ambiguous image of a penguin.} 
        \label{fig:penguin_ambiguous}
    \end{subfigure}%
    \caption{Images can be inherently ambiguous, making it necessary to quantify their uncertainty. Both images are from the ImageNet-1k benchmark dataset \citep{deng2009imagenet}.}
    \label{fig:penguins}
%\vspace{-5pt}
\end{figure}

Any human will recognize the image in \cref{fig:penguin_clear} as a penguin and store it accordingly in their mental map of animals. Now consider \cref{fig:penguin_ambiguous}. It could show a penguin, but also a seal, or maybe a beaver. The picture itself is uncertain, that is, it does not contain enough information to infer what it shows. The best any human can do is to store it somewhere in the region of aquatic animals in their mental map, flagged with a question mark.

\cref{fig:penguin_ambiguous} is no exception and the problem can not be trained away, whether we deploy a computer vision model or a human expert. \citet{beyer2020we} show that humans disagree about the class of $29.9\%$ of the images in the popular ImageNet-1k benchmark \citep{deng2009imagenet}. This is even more pronounced when images are no high-quality photographs from the internet but automatically taken by magnetic resonance imaging scanners or surveillance cameras on animal farms, as studied by \citet{schmarje2022one}. Even their best real-world dataset with only four classes to choose from has a disagreement rate of $92.2\%$. Vision inherently is and will remain ambiguous.
%Percentage of images where annotators disagree:
%ImageNet-21k: 29.9%
%/home/kirchhof/data/Benthic 0.9517156359153482
%CIFAR10H 96.81%
%/home/kirchhof/data/MiceBone 0.9726519337016575
%/home/kirchhof/data/Pig 0.922926638663671
%/home/kirchhof/data/Plankton 0.9759771986970684
%/home/kirchhof/data/QualityMRI 1.0
%/home/kirchhof/data/Synthetic 0.9333333333333333
%/home/kirchhof/data/Treeversity#1 0.9247549794498894
%/home/kirchhof/data/Treeversity#6 0.9492163647465907
%/home/kirchhof/data/Turkey 0.9670398009950248

Current deep encoders in computer vision also have mental maps, their embedding spaces, where they store what they detect in images as representation vectors. But they lack the ability to express their uncertainty: Although \cref{fig:penguin_ambiguous} is much more ambiguous than \cref{fig:penguin_clear}, both will be pinpointed to an exact representation vector in the embedding space. Any module that further processes these representations, for retrieving similar images or predicting the animal species, has no more sense of the ambiguity. 

This thesis adds uncertainty estimates to representation vectors. This enhancement may seem nuanced at first glance, but it conceals an iceberg of challenges beneath the surface: Representations are latent variables, meaning that they are not observable in the real world and a computer vision model has to find them itself. Adding uncertainties, which are also not observable in the real world, to such already unobservable latent variables makes the problem even more complicated. And more still, we also have to compare these unobservables about unobservables against some notion of observable ground-truth in the real world to ensure their quality and good performance, which is a territory uncharted by state-of-the-art uncertainty benchmarks. But the reward of solving these challenges is high: Uncertainties added directly to representations will trickle down to all applications that start off from representations, e.g., of pretrained models.

Before diving into details, we give a teaser of our results in \cref{fig:clustering}. This is the embedding space of six dog breeds, where more uncertain representations are larger and more transparent. The plot makes it easy to understand which images are more ambiguous and likely to be misclassified because they reveal too few information. This enables computer vision models to automatically treat uncertain images differently, e.g., by abstaining from classifying them until a user inputs a more clear image, enabling a more trustworthy deployment of machine learning models in practice \citep{mucsanyi2023trustworthy}. Notably, these uncertainties are provided zero-shot. The model was trained on a different dataset and the uncertainties are generalized from the large pretraining corpora we scaled our approach to. This means that practitioners can make direct use of the findings of this thesis by downloading our pretrained uncertainty models and running them on their data. 

Let us now return to the start of our expedition, formally defining representation learning and uncertainties in computer vision as well as our precise research questions, which culminated in the development of these general-purpose representation uncertainties.

\section{Related Work}

\subsection{Pretrained Representations}

\afterpage{%
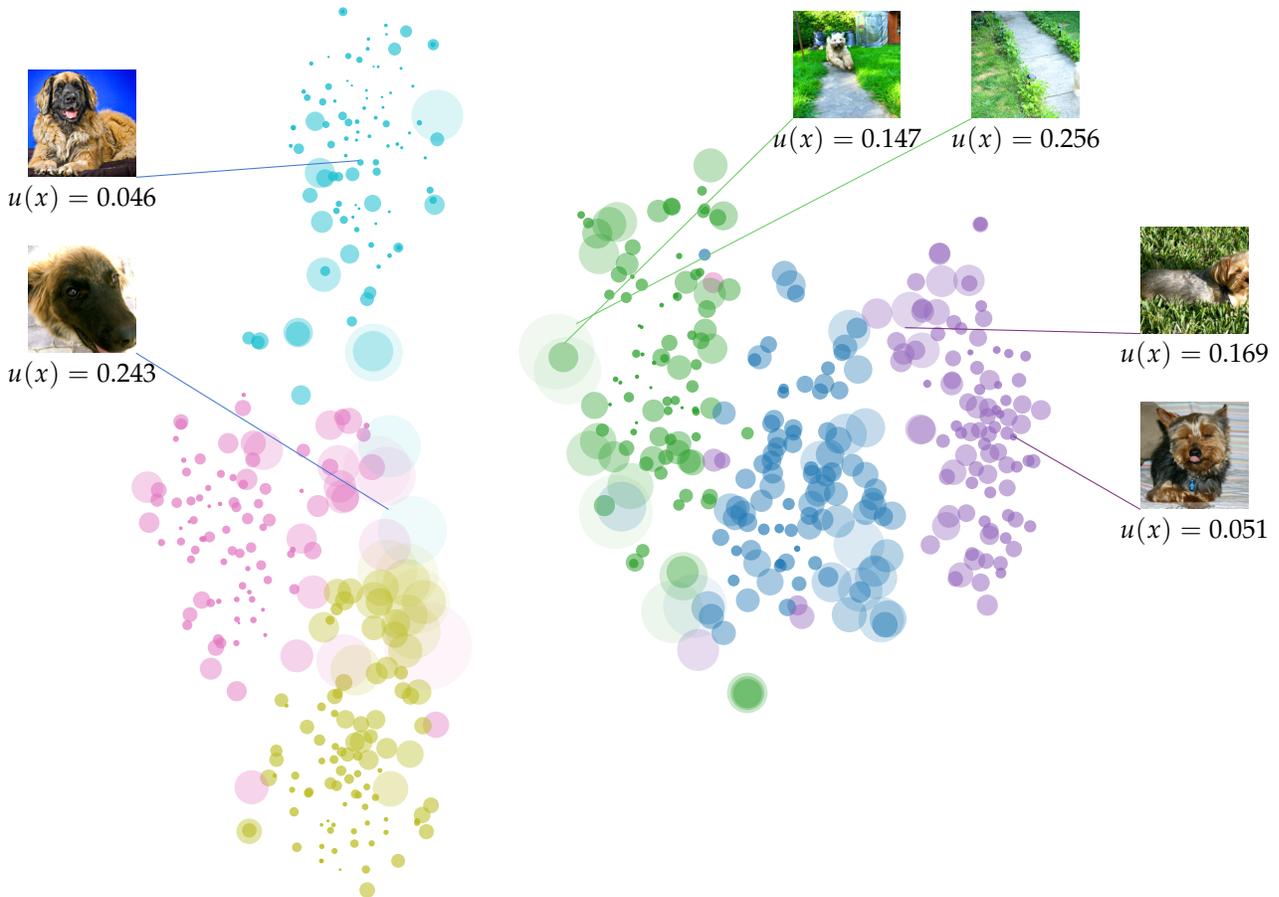
\begin{figure}[ht]
    \centering \vspace{-7mm}
    \begin{tikzpicture}
    \node[anchor=south west,inner sep=0] (image) at (0,0) {\includegraphics[width=0.795\linewidth]{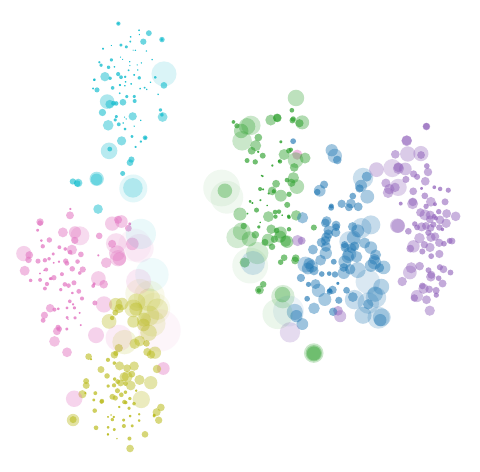}};
    \begin{scope}[x={(image.south east)},y={(image.north west)}]
        %\draw[help lines,xstep=.01,ystep=.01] (0,0) grid (1,1);
        %\draw[help lines,xstep=.1,ystep=0.1,color=black] (0,0) grid (1,1);
        %\node[anchor=north west,inner sep=0] (image1) at (0.45,0.18) {\includegraphics[scale=0.2]{figures/small_bottom.png}};
        \node[anchor=north west,inner sep=0] (image1) at (-0.07,0.71) {\includegraphics[scale=0.25]{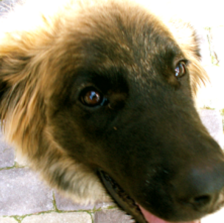}};
        \node[below=1mm of image1,inner sep=0] (text1) {\footnotesize$u(x)=0.243$};
        \draw[blue] (image1.south east) -- (0.293, 0.44);
        \node[anchor=north west,inner sep=0] (image2) at (-0.07,0.89) {\includegraphics[scale=0.25]{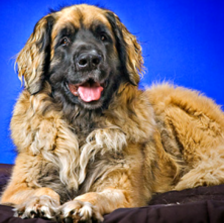}};
        \node[below=1mm of image2,inner sep=0] (text2) {\footnotesize$u(x)=0.046$};
        \draw[blue] (image2.south east) -- (0.263, 0.797);
        \node[anchor=north west,inner sep=0] (image3) at (0.7,0.95) {\includegraphics[scale=0.25]{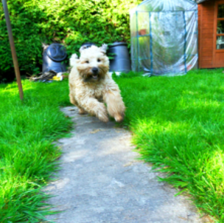}};
        \node[below=1mm of image3,inner sep=0] (text3) {\footnotesize$u(x)=0.147$};
        \draw[green] (image3.south west) -- (0.47, 0.61);
        \node[anchor=north west,inner sep=0] (image4) at (0.88,0.95) {\includegraphics[scale=0.25]{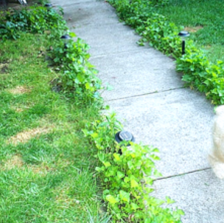}};
        \node[below=1mm of image4,inner sep=0] (text4) {\footnotesize$u(x)=0.256$};
        \draw[green] (image4.south west) -- (0.482, 0.63);
        \node[anchor=north west,inner sep=0] (image5) at (1.05,0.73) {\includegraphics[scale=0.25]{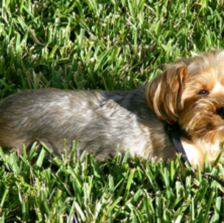}};
        \node[below=1mm of image5,inner sep=0] (text5) {\footnotesize$u(x)=0.169$};
        \draw[purple] (image5.south west) -- (0.813, 0.626);
        \node[anchor=north west,inner sep=0] (image6) at (1.05,0.55) {\includegraphics[scale=0.25]{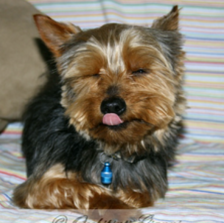}};
        \node[below=1mm of image6,inner sep=0] (text6) {\footnotesize $u(x)=0.051$};
        \draw[purple] (image6.south west) -- (0.922, 0.515);
        %\node[anchor=north east,inner sep=0] (image7) at (-0.05,0.55) {\includegraphics[scale=0.25]{figures/white.png}}; %this one is just to make the plot centered, to "balance out" the images on the right
    \end{scope}
    \end{tikzpicture} \vspace{-8mm}
    \caption{Each point is the representation of an Oxford Pets image \citep{parkhi12a}. The size of each dot visualizes the uncertainty $u(x)$ of the representation, calculated by our approach in \cref{chap:icml2}. This makes it easier to detect images that are naturally ambiguous (large and transparent). Figure cited from the original paper \citep{kirchhof2024pretrained}.}
    \label{fig:clustering}
\end{figure}
}

We first establish notation that will reoccur throughout the thesis. Let $x$ denote an input from the input space $\mathcal{X}$ and $y$ an output from the output space $\mathcal{Y}$. The task of any machine learning model is to fit a function $f:\mathcal{X} \rightarrow \mathcal{Y}$ that predicts $y$ from $x$. In computer vision, these are commonly an image $x$ and a class label $y$. To predict a label from an unstructured information source like an image, modern deep learning architectures first extract features from the image. That is, they have an encoder component $e:\mathcal{X} \rightarrow \mathcal{Z}$ and a subsequent classifier $c:\mathcal{Z} \rightarrow \mathcal{Y}$ with $f = c \circ e$. The goal of representation learning \citep{bengio2013representation} is to learn a versatile encoder $e$ such that the representations $z$ in the embedding spaces $\mathcal{Z}$ in the middle represent the content of the image in an abstract manner, as a high dimensional latent vector.

Since training the encoder is often the most time and data consuming part of training a deep learning architecture, the representations are often trained in advance on large image datasets with versatile classes to learn diverse features. Most modern computer vision projects start off from such pretrained models. Pretrained representations enable, e.g., to search for semantically similar images, called retrieval \citep{chang2005relational,liu2021image,jush2023medical,douze2024faiss}, or to re-use the encoder and learn new tasks quicker by finetuning on a small dataset in a few-shot or even zero-shot manner \citep{weiss2016survey,goyal2023finetune,ramesh2021zero}.  

The key to well usable pretrained representations is that semantically similar images should lay close to one another in the embedding space. Contrastive learning approaches \citep{chopra2005learning,semihard,contrastive,chen2020simple,grill2020bootstrap,radford2021learning,chen2021exploring} formulate this directly as an objective function when training the encoder. As an example, tuples of images cropped from the same underlying image are encouraged to be close to one another \citep{contrastive} or they are encouraged to be close to one another and far from other images \citep{semihard}. Besides these self-supervised approaches, modern pretraining approaches also return to traditional supervised learning, where the supervision signal that tells if two images are similar is whether they have the same class label. The key here is that the class labels are diverse enough, such as on the ImageNet-1k dataset \citep{deng2009imagenet} that comprises 1.2 million images of 1000 classes or on ImageNet-21k \citep{deng2009imagenet} that comprises 14.2 million images of 21.8 thousand classes. 

These advances in pretrained representations reveal important desiderata for our desired representation uncertainties: If representation uncertainties are output along with pretrained representations, we need to pretrain them on similar scales. We also need to ensure that our representation uncertainties capture uncertainties of the general image content that the representation summarizes, not just uncertainty in terms of the (pre-) training task. With these properties in mind, let us review the current state uncertainty estimation approaches in computer vision.

\subsection{Large-scale Uncertainties in Computer Vision}

Uncertainty estimation adds a second task to the model. In addition to the estimate for $y$ it also has to output an uncertainty estimate $u(x), \,u: \mathcal{X} \rightarrow \mathcal{U}$, sometimes in the form of a probability $\mathcal{U} = [0,1]$ or more generally any scalar value with $\mathcal{U} \subseteq \mathbb{R}$. These uncertainty estimates are a key prerequisite to deploy models in safety-critical areas like medical imaging or self-driving cars \citep{gulshan2016development,carannante2021trustworthy,kurz2022uncertainty,franchi2022muad} where we want to predict only if we are certain. Uncertainties are also fundamental ingredients of anomaly detection \citep{chalapathy2019deep} and active learning \citep{settles2009active,nguyen2022measure}.

First attempts to bring uncertainty into deep learning and computer vision stem from Bayesian roots \citep{bernardo2009bayesian}. A prominent example is the Laplace approximation \citep{mackay1992bayesian} that approximates a Gaussian around the network's parameters. This allows to sample multiple output vectors per input, which can be processed into scalar uncertainty estimates $u(x)$. The issue is that these approximate Bayesian approaches are hard to scale to deep architectures, with work on scaling going on to this date \citep{ritter2018scalable,daxberger2021laplace,deng2022accelerated}. Later works thus yield these samples more directly: Deep ensembles \citep{lakshminarayanan2017simple} train multiple networks to output multiple vectors, and \citet{gal2016dropout} activate random Dropout \citep{srivastava2014dropout} at inference time to produce multiple slightly different outputs. These approaches are scalable to arbitrarily deep architectures, but their runtime (and memory) costs still scale linearly in the number of samples. Anyhow, they mark the start of a trend: Reducing the computational hurdle of uncertainty estimation to enable a widespread application in practice.  

A recent step towards low-cost uncertainties is to output uncertainty estimates $u(x)$ directly during the forward pass of the model, so called deterministic methods \citep{pmlr-v162-postels22a,haussmann2020sampling}. This is implemented by adding modules to the architecture that output specialized uncertainties $u(x)$. These modules are trained for specialized tasks \citep{mucsanyi2024benchmarking}. For example, to detect out-of-distribution inputs \citep{galil2023a}, deterministic methods estimate the data density in the model's latent space \citep{lee2018simple,vanamersfoort2020uncertainty,mukhoti2023deep}. To estimate correctness of prediction, they predict the model's own loss at any given sample \citep{Yoo_2019_CVPR,cui2023learning,lahlou2023deup,laves2020well}. Such specialized approaches are also relevant for the recent strive for decomposed or disentangled uncertainties \citep{wimmer2023quantifying,pmlr-v202-bengs23a,gruber2023sources,valdenegro2022deeper,depeweg2018decomposition}. We contribute to these disentanglement efforts in \cref{chap:icml2}.

So how could one build a specialized uncertainty estimate for representations? One strain of research are probabilistic embeddings \citep{oh2018modeling,collier2023massively,pmlr-v202-kim23g,nakamura2023representation}. They add an auxiliary output head that estimates a variance parameter for each representation, resulting in a distribution over all possible latents that an input could show. While probabilistic embeddings have shown increased performance \citep{karpukhin2022probabilistic} and qualitatively sensible outputs \citep{oh2018modeling,scott2019stochastic}, their theoretical underpinning and evaluation metrics are still in their infancy. We contribute to these efforts in \cref{chap:eccv,chap:icml,chap:neurips}.

The last gap to bridge uncertainty estimates with pretrained representations is their transferability. Initial works \citep{guo2017calibration} but also current large-scale undertakings \citep{dehghani2023scaling} often consider calibration only on the dataset the model was trained on. The largest distribution shifts that uncertainties are evaluated on are corrupted versions of the train dataset \citep{ovadia2019can,tran2022plex}. At a certain level of corruption, let alone on new datasets with new classes, images are commonly considered out-of-distribution, so that the evaluation protocol is just to achieve high uncertainties on these samples compared to in-distribution ones \citep{park2023understanding,galil2023a,pmlr-v162-postels22a,ovadia2019can}. This is a reasonable goal, but we want to take the generalization of uncertainty estimators one step further. Similar to representation learning, we want uncertainties to work within completely new datasets, where uncertainties should not be generally high but differ between the unseen samples. Also similar to representation learning, this requires novel benchmarking metrics, see \cref{chap:neurips}, and large-scale pretrained models developed along these metrics, see \cref{chap:icml2}.

\section{Research Questions}

The main goal of this thesis is to provide uncertainty estimates along with representation vectors, merging the current enhancements in representation learning with those in uncertainty estimation. As we have outlined before, this is intended to make uncertainties transferable to new datasets and tasks and thus easier to use. To this end, we need to develop new approaches, gain a theoretical understanding about representation uncertainties and how to benchmark them, and finally scale our findings to match current pretrained representation models. We take on these challenges one by one in the following chapters, guided by the following research questions (RQs).

\begin{enumerate}[label={RQ\arabic*.},leftmargin=5em]
    \item Which methods can provide uncertainties about representations?
    \item What do these uncertainties reflect theoretically or in the real-world?
    \item How to quantify how good an uncertainty about a representation is?
    \item Can we scale the uncertainties to large models and large pretraining corpora?
\end{enumerate}

\section{Contributions and Outline}

We answer these questions in four chapters, each resembling one paper. We summarize their main findings and contributions to the scientific community below.

\cref{chap:eccv} takes on RQ1, providing first methods that add uncertainties to representations in a subfield of representation learning, deep metric learning. In particular, we utilize the probabilistic embeddings framework where we predict a mean and a variance for each representation. We expand these works by implementing distribution-to-distribution distance functions and generalizing previous distributions to more flexible covariance structures. These enhancements make deep metric learning probabilistic and we find that they increase the performance. We take a first look at RQ2 to analyze why the uncertainties lead to higher performance.

\cref{chap:icml} provides deeper theoretical insights into RQ2. We derive a theoretical framework to define representation uncertainties formally and to investigate in which sense they are correct. We find that they can be seen as the posteriors of a lossy image-generating process, generalizing previous theoretical results of nonlinear independent component analysis \citep{dml_inversion,reizinger2022jacobian}. We also provide a new method to learn representation uncertainties in self-supervised, adding to RQ1. We verify that its representation uncertainties are indeed correct in the theoretical sense and, in the practical sense of RQ3, correlated with human uncertainties. 

\cref{chap:neurips} provides the community with the first benchmark to measure correctness of representation uncertainties at scale and under distribution shifts, answering RQ3. This allows representation learning researchers to enhance their benchmarking code with metrics for uncertainties in four lines of code. We verify our benchmark with comparisons to the human and other real-world uncertainties studied in the previous chapter to broaden our understanding of RQ2. We reimplement both of our approaches from RQ1 and add multiple ones from literature. We scale them all to ImageNet-1k and use a pretraining-like train and test framework, paving the way for RQ4. 

\cref{chap:icml2} uses this benchmark to develop pretrained representation uncertainties at the scale of ImageNet-21k and large Vision Transformer, aiming for RQ4. This compiles the findings of all previous chapters and research questions into one downloadable model for future researchers. In the process, we fix a gradient conflict that deteriorated the performance of deterministic uncertainty approaches in the literature. We also study the behaviour of our uncertainties and find that they provide aleatoric uncertainties devoid from epistemic uncertainties, contributing to RQ2 and to the recent efforts in uncertainty disentanglement \citep{wimmer2023quantifying,mucsanyi2024benchmarking}.

Finally, in \cref{chap:conclusion} we outline the applications that representation uncertainties enable and discuss how our findings shape future directions in uncertainty quantification.

\section{List of Publications}

\subsection{Publications Relevant to this Thesis}

This thesis comprises four main publications that I have published as first author in the last three years. All papers are summarized and discussed in their corresponding chapters in the main text, and appended to the thesis in their full form.

\begin{enumerate}[label={},leftmargin=3.3em,rightmargin=3.3em]
\item \textbf{Michael Kirchhof}, Karsten Roth, Zeynep Akata, and Enkelejda Kasneci. A non-isotropic probabilistic take on proxy-based deep metric learning. \textit{European Conference on Computer Vision (ECCV)}, 2022. 
\item \textbf{Michael Kirchhof}, Enkelejda Kasneci, and Seong Joon Oh. Probabilistic contrastive learning recovers the correct aleatoric uncertainty of ambiguous inputs. \textit{International Conference on Machine Learning (ICML)}, 2023. 
\item \textbf{Michael Kirchhof}, Bálint Mucsányi, Seong Joon Oh, and Enkelejda Kasneci. URL: A representation learning benchmark for transferable uncertainty estimates. \textit{Neural Information Processing Systems Track on Datasets and Benchmarks (NeurIPS D\&B)}, 2023.
\item \textbf{Michael Kirchhof}, Mark Collier, Seong Joon Oh, and Enkelejda Kasneci. Pretrained visual
uncertainties. arXiv preprint arXiv:2402.16569, 2024. % Under submission.
\end{enumerate}

\subsection{Further Publications}

Besides these main works, I have been involved in three major side projects during my time as a Ph.D. student, all centered around representations and uncertainty. 

\begin{enumerate}[label={},leftmargin=3.3em,rightmargin=3.3em]
\item Tobias Leemann, \textbf{Michael Kirchhof}, Yao Rong, Enkelejda Kasneci, and Gjergji Kasneci. When are post-hoc conceptual explanations identifiable? \textit{Uncertainty in Artificial Intelligence (UAI)}, 2023.
\item Bálint Mucsányi, \textbf{Michael Kirchhof}, Elisa Nguyen, Alexander Rubinstein, and Seong Joon Oh. Trustworthy machine learning. arXiv preprint arXiv:2310.08215, 2023.
\item Bálint Mucsányi, \textbf{Michael Kirchhof}, and Seong Joon Oh. Benchmarking uncertainty disentanglement: Specialized uncertainties for specialized tasks. arXiv preprint arXiv:2402.19460, 2024. Under submission.
\end{enumerate}

%% file: paper1.tex
\chapter{Probabilistic Representation Learning}
\label{chap:eccv}

\input{paper1/main}

%% file: paper1/main.tex
\textbf{Michael Kirchhof}, Karsten Roth, Zeynep Akata, and Enkelejda Kasneci. A non-isotropic probabilistic take on proxy-based deep metric learning. \textit{European Conference on Computer Vision (ECCV)}, 2022. 

\section{Prologue}

% A week before christmas...

% gathering information, as a Bayesian coming slowly to the conclusion that typical embeddings cannot represent uncertainties. If at all, they would have to learn them in-and-out, mapping deteriorated versions of an image to individual positions in the latent space. So I tried out a recent technique, probabilistic embeddings, and started working on distance functions to compare them. 

"That probabilistic approach seems to work, I'm right now at 66\% Recall@1 on CUB.", I wrote to Karsten Roth, at that time a Ph.D. student specializing in representation learning. I had just sent him an initial implementation of what happens if we change vector representations to distributional or probabilistic ones. What had happened was that it outperformed most baselines in his recent benchmark. He responded within seconds, with a scientist's mixture of excitation and caution, "I'd have time for a meeting later today? Let's just double check the code.". The code was fine and by evening we had set up our collaboration. In combining our strenghts, we expanded the initial idea mathematically and gained more evidence empirically. We presented the results at a computer vision conference, ECCV, that would have impacts on the next chapters of this dissertation.

\section{Motivation}

We develop our uncertainties for representations from classical representation learning. The goal here is to encode images into vectors, such that images of, e.g., the same class, have similar vector representations. This is a mandatory property for retrieval systems \citep{sohn2016improved,brattoli2020rethinking,douze2024faiss}. One sub-field of this is deep metric learning \citep{roth2020revisiting}. It investigates which distance function between the representations one should use to train the encoder. We find that simple ones like cosine distance do not account for uncertainties in the images, despite the field having argued that this was an intended feature to ensure all images were treated the same \citep{ranjan2017l2}. We, along with concurrent works \citep{scott2021mises}, question this and argue that uncertainties are informative features that support the training. In this chapter, we represent images as distributions over possible latents instead of single vectors, so called probabilistic embeddings. We show how to calculate distances between them, and how much and why this improves deep metric learning.

\section{Methods}

\begin{figure}[t]
    \centering
    %\captionsetup[subfigure]{oneside,margin={0.8cm,0cm}}
    \begin{subfigure}{0.47\textwidth}
    \centering
        \includegraphics[width=0.7\linewidth, trim={5.9cm 3.4cm 5.2cm 3.4cm}, clip]{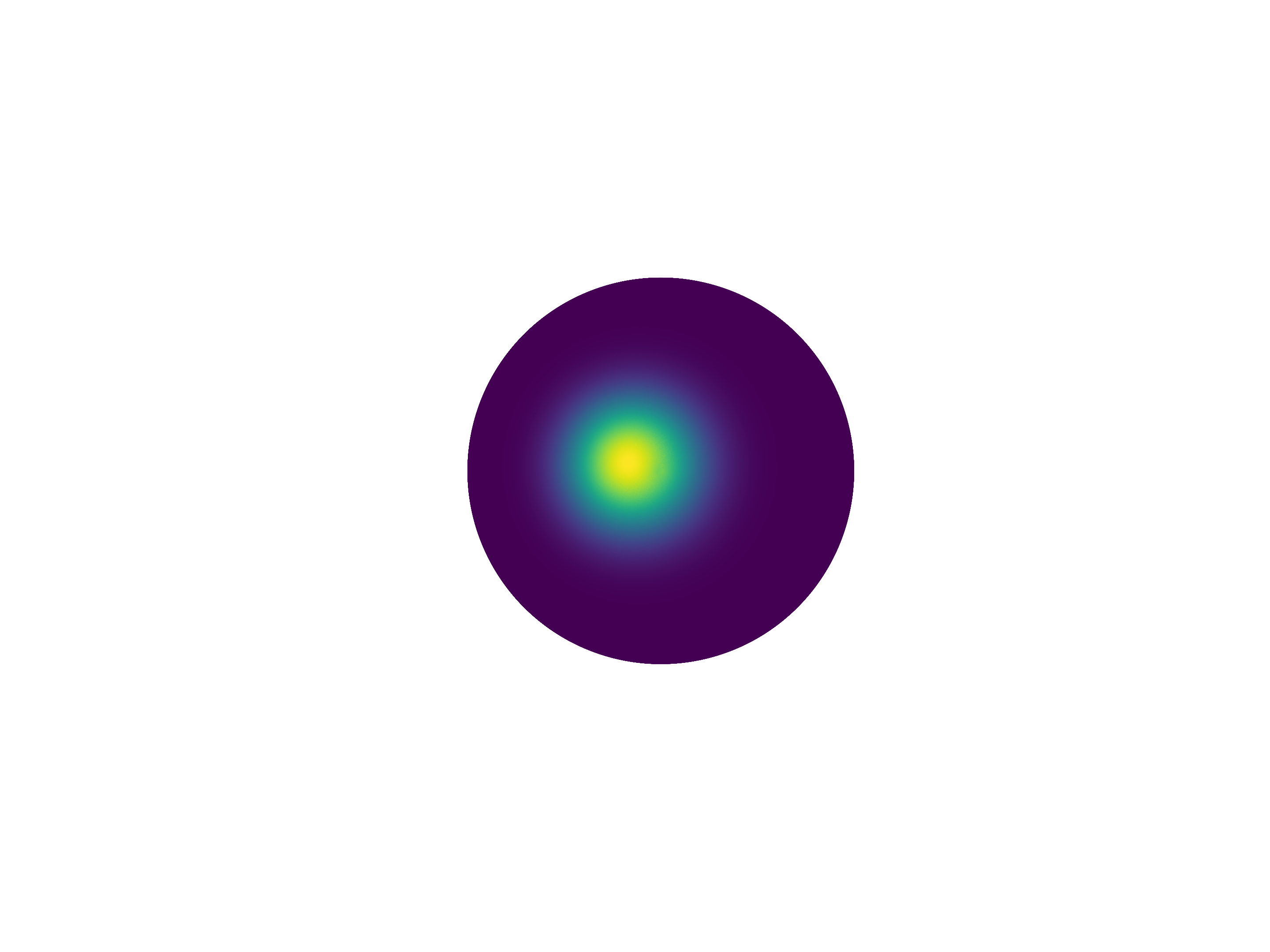}
        \caption{von Mises Fisher (vMF)} 
        \label{fig:nivmfa}
    \end{subfigure}%
    \hfill{\hbox{}}   % maximize separation between the subfigures
    \begin{subfigure}{0.47\textwidth}
        \centering
        \includegraphics[width=0.7\linewidth, trim={5.9cm 3.4cm 5.2cm 3.4cm}, clip]{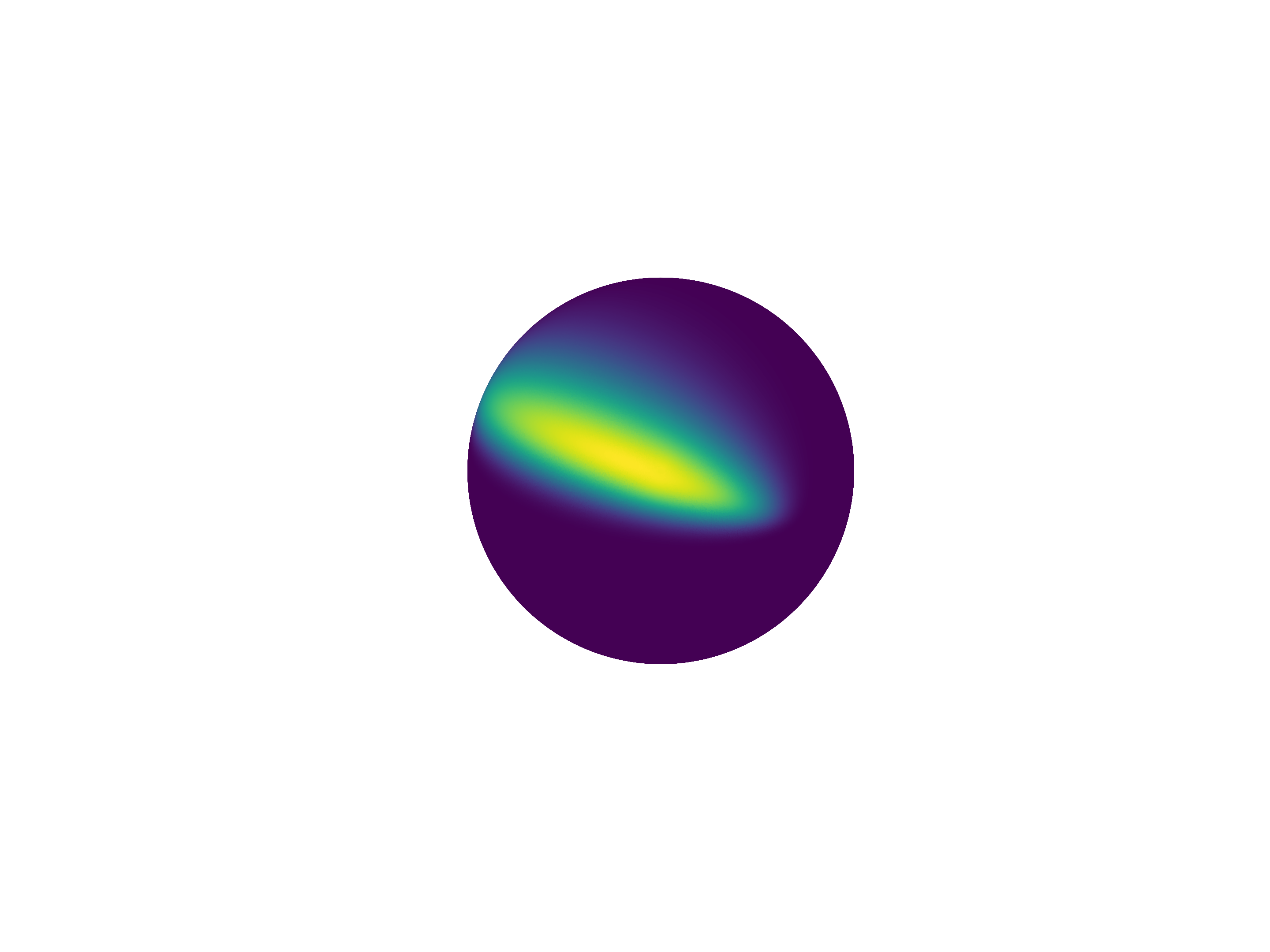}
        \caption{non-isotropic von Mises-Fisher (nivMF)} 
        \label{fig:nivmfb}
    \end{subfigure}%
    \caption{Densities of a vMF and a non-isotropic vMF distributions on a three-dimensional unit-sphere. Purple is a low and yellow a high density. Figure adapted from the original paper \citep{kirchhof2022non}.}
    \label{fig:nivmf}
%\vspace{-5pt}
\end{figure}

The goal of deep metric learning is to learn representations $e(x)$ for each image such that similar images are placed close to one another and dissimilar ones far from another in a model's representation space $\mathcal{Z}$. Similarity is usually defined as belonging to the same class or being two crops from the same image. These representations are learned by loss functions that measure the distance between representations and push similar ones closer to one another and dissimilar ones away from each other. To reduce the noise in this process, ProxyNCA and ProxyNCA++ \citep{proxynca,teh2020proxynca} propose to use proxies for each class, so that each image is pushed closer to the proxy of its class. The contrastive loss function is %The outer part of this objective is a simple softmax but the inner part uses a distance to gauge the distance between a sample and a proxy. This distance is the key to learning a well-structured embedding space. The default is to use a simple cosine distance, assuming a unit-sphere as an embedding space.
\begin{equation}
\label{eq:pnca}
    \mathcal{L}_\text{NCA++} =  \log\frac{\exp\left(s\left(\frac{e(x)}{\lVert e(x) \rVert}, \text{p}^*\right) / t\right)}{\sum_{c=1}^C \exp\left(s\left(\frac{e(x)}{\lVert e(x) \rVert}, \text{p}_c\right) / t\right)} \,,
\end{equation}
where $\text{p}^*$ is the true proxy (i.e., class) of $x$, $t > 0$ is a temperature, and $p_c, c=1,\dotsc, C,$ are all $C$ possible classes. In practice, the representations $e(x)$ are often normalized to unit length, so the representation space $\mathcal{Z}$ is a unit sphere and the similarity function $s$ is a cosine similarity. 

Our key idea is to allow uncertainties about what an image represents, e.g., if it is blurry or an information-losing crop. For this, we represent images as distributions $\zeta(x)$ over all possible latents, so called probabilistic embeddings. In particular, we use von Mises-Fisher (vMF) distributions \citep{fisher1953dispersion,mardia2009directional} over the unit-sphere $\mathcal{Z}$, as shown in \cref{fig:nivmf}. For proxies, we develop non-isotropic von Mises-Fisher (nivMF) distributions $\rho$. They allow class distributions to have non-unit covariances. To measure the distribution-to-distribution similarity between probabilistic embeddings and proxies, we use the expected likelihood kernel (ELK, \citeauthor{jebara2003bhattacharyya}, \citeyear{jebara2003bhattacharyya}). These changes result in the uncertainty-aware contrastive loss function
\begin{equation}
\label{eq:pnca_distr}
    \mathcal{L}_{\text{nivMF}} = \log \frac{\exp(s(\zeta(x), \rho^*) / t)}{\sum_{c=1}^C \exp(s(\zeta(x), \rho_c) / t)} .
\end{equation}

We implement this by using auxiliary learnable variables for the proxy means and covariances. For the probabilistic embeddings of images, we use the typical (normalized) representations $\frac{e(x)}{\lVert e(x) \rVert}$ as the mean value of the $\zeta(x)$ distribution. The concentration (inverse variance) parameter is set to the pre-normalization representation norms, i.e., $\lVert e(x) \rVert$, following \citet{li2021spherical}. This utilizes that the representation norm is empirically related to certainty, namely how many class characteristic features can be detected in an image. We dicuss this further in the main paper.

\begin{figure}[t]
    \centering
    \begin{subfigure}{0.47\textwidth}
    \centering
        \includegraphics[width=\linewidth]{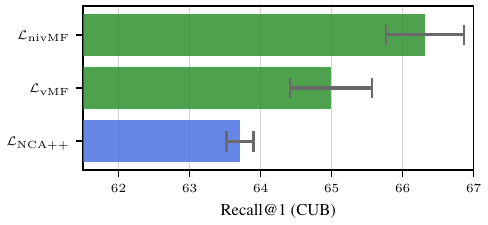}
        %\caption{CUB} 
        %\label{fig:abl_cub}
    \end{subfigure}% % maximize separation between the subfigures
    \begin{subfigure}{0.47\textwidth}
        \centering
        \includegraphics[width=\linewidth]{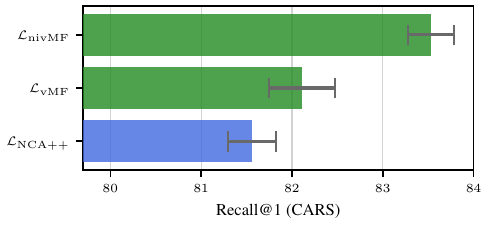}
        %\caption{CARS} 
        %\label{fig:abl_cars}
    \end{subfigure}%
    \caption{Probabilistic embeddings (green) lead to better retrieval performance than deterministic ones (blue). Bars show the standard deviation across five seeds. Figure adapted from the original paper \citep{kirchhof2022non}.}
    \label{fig:abl}
\end{figure}

\section{Core Results}

\subsection{Probabilistic Embeddings Improve Retrieval Performance}

The primary objective of deep metric learning is to learn a well-structured embedding space. Representations of similar images should be close to one another, enabling retrieval. This is measured via the Recall@1: If we compute mean representations for each image in the test dataset, how often is the nearest neighbor of each representation in the same class? This percentage is computed on the dataset the model was trained on, but on a withheld set of classes. This induces a small domain shift to ensure the representations generalize beyond the training classes.

\cref{fig:abl} shows that training probabilistic embeddings with $\mathcal{L}_{\text{nivMF}}$ leads to a higher Recall@1 than the vector representations of $\mathcal{L}_{\text{NCA++}}$ on two representation learning benchmark datasets, CUB-200 \citep{cub200} and CARS-196 \citep{cars196}. As a step in-between, $\mathcal{L}_{\text{vMF}}$ uses vMF distributions for both images and classes, showing that both enhancements, probabilistic embeddings and non-isotropy, increase retrieval performance. The main paper presents similar advantages when adding probabilistic embeddings to more complicated contrastive loss functions.

\subsection{Uncertainties Re-weight the Gradients}

The above experiment does not use the learned uncertainties during testing. It only measures the Recall@1, i.e., nearest (mean) representation in terms of cosine similarity. So training with uncertainties has helped learn better representations, but how? 

In the main paper, we analyze our probabilistic loss analytically. We find that the gradient that each image has on the representations is scaled up by its certainty. More certain images receive a higher weight during training than uncertain images. This reduces the impact of samples that are low-quality or potentially mislabeled. We provide more details and comparisons to the deterministic $\mathcal{L}_{\text{NCA++}}$ in the main paper in the appendix.

\section{Discussion}

This work centered around RQ1, finding a method to add uncertainties to representations, namely probabilistic embeddings. Our main finding is that uncertainties are not just an end unto themselves, but help learn better representations. This work also contributed fundamentals that we will see reoccurring in the next chapters, such as the non-isotropic vMF distribution or a corrected approximation to the vMF normalizing constant, excluded here but detailed in the main paper. We have also touched upon RQ2, gaining a first understanding of how uncertainties benefit representation learning.

But the main limitation is that we have evaluated the learned uncertainties only indirectly. They helped learn representations with higher retrieval performance, but we have not evaluated how correct the uncertainties are in and by themselves. In fact, one could argue that they are also trained only indirectly, since they are parametrized by the representation vector norms which other side effects during training could have influenced. We investigate these two open points in the next chapter to verify and expand our understanding of representation uncertainties.

% This explorative work focussed on the advantage for the main task, less on the uncertainties themselves.

% Did first investigations on the learned uncertainties themselves, but will focus more on them next

% norms instead of explicitly learning them

%% file: paper2.tex
\chapter{Probabilistic Embeddings are Provably Correct}
\label{chap:icml}

\input{paper2/main}

%% file: paper2/main.tex
\textbf{Michael Kirchhof}, Enkelejda Kasneci, and Seong Joon Oh. Probabilistic contrastive learning recovers the correct aleatoric uncertainty of ambiguous inputs. \textit{International Conference on Machine Learning (ICML)}, 2023. 

\section{Prologue}

"But how do you know that these variances are \emph{correct}?". It was at a poster session at ECCV where I presented the previous paper, and I had just encountered a mind-reader (or my mind was very easy to read). This question turned out to be not just in my mind, but to be haunting the field ever since probabilistic embeddings, or even variational auto-encoders, were invented. In parallel, a new researcher came to Tübingen: Seong Joon Oh. I knew his name. "Aren't you the author of hedged instance embeddings, the first paper on probabilistic embeddings?", I asked. He confirmed. And he also confirmed that he had also been looking for a mathematical answer to the upper question. We compared our notes and so started hours-long discussions of potential proof techniques, thought experiments, and scrutiny of potential loopholes. My coworkers may remember me sitting in the office for days, weeks, and months on end without a laptop, only with countless scribbled papers and a pencil. We succeeded eventually, and the chapter below summarizes our mathematical formalization of the question, as well as its answer.

\section{Motivation}

The previous chapter introduced probabilistic embeddings as a way to represent uncertainties in representation spaces. And they indeed work in the sense that they improve performance. But what is it that their variance parameters capture? Are they indeed the \textit{correct} uncertainties (and if yes, in which sense)? To establish a ground for mathematical arguments, we first need a formal framework. We generalize the non-linear independent component analysis framework of \citet{dml_inversion} to formalize data-generating processes that lose information while generating images, where the amount of lost information equals the uncertainty that the probabilistic embeddings have to resemble. The challenge here is that the amount of lost information is also a lost information -- we only have access to the final image without any further supervision on how uncertain it is. Strikingly, we find a loss function called MCInfoNCE whose probabilistic embeddings are provably correct: Their variances exactly reflect the amount of lost information, while being learned solely from self-supervision. 
\begin{figure}[t]
    \centering \vspace{-5mm}
    \includegraphics[width=0.95\textwidth]{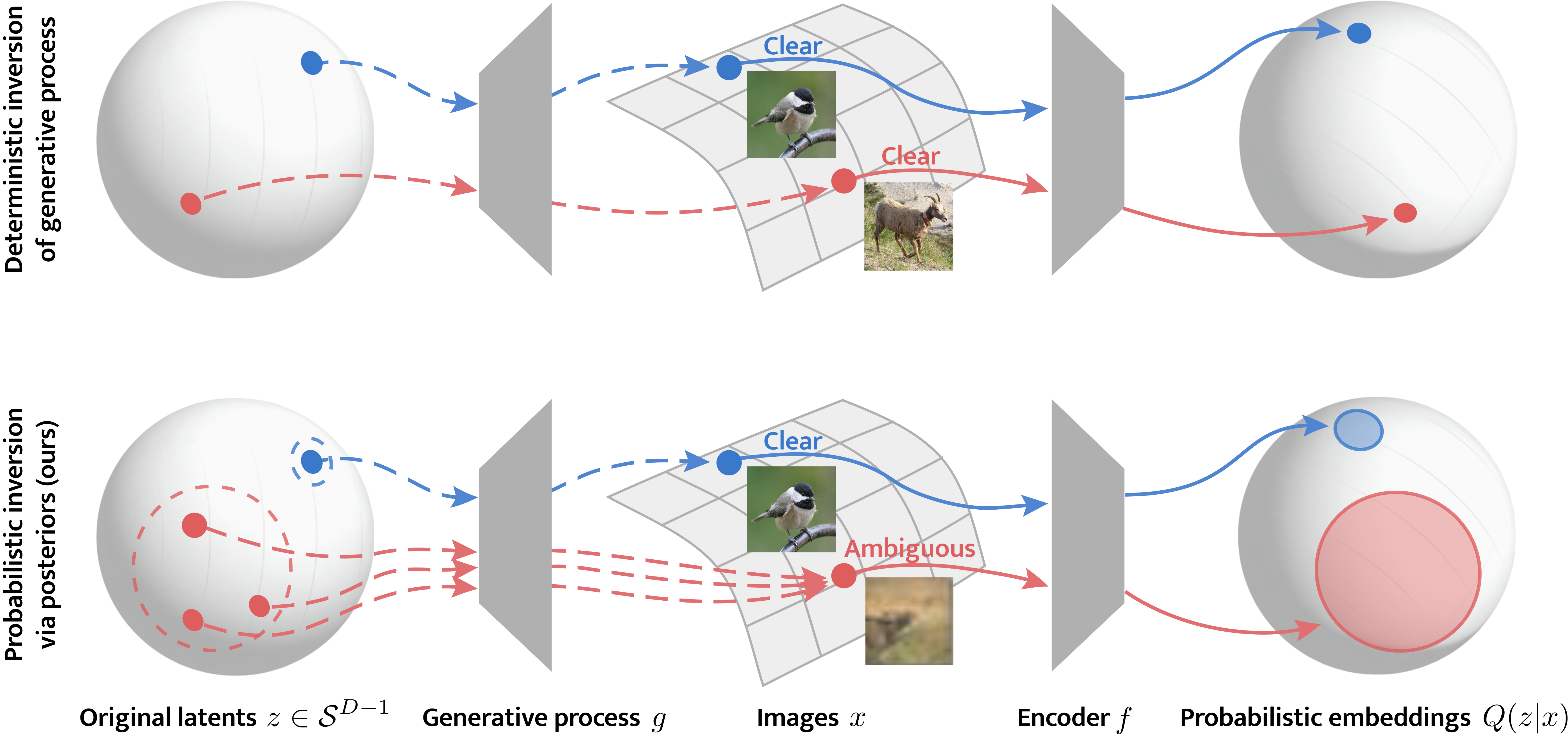}
    \caption{Images are created from unknown latent vectors by a data-generating process. Deterministic image representations intend to rediscover this vector (top). When the data-generating process is probabilistic (bottom) and creates ambiguous images, it loses information about the latent vectors, so that several ones could have created the image. Probabilistic embeddings recover this posterior, which we prove for MCInfoNCE. Figure cited from the original paper \citep{kirchhof2023probabilistic}.}
    \label{fig:overview}
\end{figure}

\vspace{-2mm}\section{Methods}

To be able to discuss any notion of correctness, let us first formalize how images are generated. Following \citet{dml_inversion}, we assume that some data-generating process turns latent vectors into images, as depicted in \cref{fig:overview}. In mathematical terms, a function $g: \mathcal{Z} \rightarrow \mathcal{X}$ maps the latent $z$ to an image $x$. To add uncertainties to the process, this mapping is not deterministic. The same latent could be mapped to different images, blurred, cropped, or pixelated in different ways. In statistical terms, $g$ is no more a function with one output per input, but a likelihood $P(X|Z)$. This likelihood is even more complicated than the already complicated generative process $g$. The trick is that we are not actually interested in the generator $P(X|Z)$ -- we are only interested in reconstructing $z$ from $x$, i.e., in its posterior $P(Z|X)$. This posterior describes which latents $z$ could have generated the image $x$. If more information about $z$ is lost during the generation of $x$ and $x$ could match several possible $z$, the posterior becomes wider. 

This is what probabilistic embeddings $Q(Z|X)$ are designed to represent. Consequently, we can define correctness for them: Probabilistic embeddings $Q(Z|X)$ (including their uncertainty parameters or variances) are \textit{correct} iff they are equal to the posteriors $P(Z|X)$ of the generative process. What is left to show is that some loss function is minimized by probabilistic embedding estimates only if they are equal to the true posteriors.

To this end, we introduce the MCInfoNCE loss
\begin{align}
    \mathcal{L}_\text{MCInfoNCE} = -\log \mathop{\mathbb{E}}\limits_{z \sim Q(z|x)} \mathop{\mathbb{E}}\limits_{z^+ \sim Q(z^+|x^+)} \mathop{\mathbb{E}}\limits_{z^-_m \sim Q(z^-_m|x^-_m)} \left( \frac{e^{\kappa_\text{pos} z^\top z^+}}{\frac{1}{M} e^{\kappa_\text{pos} z^\top z^+} + \frac{1}{M} \sum\limits_{m=1}^M e^{\kappa_\text{pos} z^\top z^-_m}} \right)\,\,.
\end{align}
The innermost part is a self-supervised InfoNCE loss \citep{oord2018representation} that trains the representation $z$ of an image to be closer to a positive partner $z^+$ (a crop of the same image) than to negatives $z^-_m$ (other images in the batch). This whole inner term is then evaluated not over predicted deterministic representations $z$ but over representations $z$ drawn from the predicted probabilistic embeddings, usually 4 to 16 samples. We implement the probabilistic embeddings by vMF distributions whose variances are learned by an MLP head. This adjustment to turn InfoNCE into the probabilistic MCInfoNCE is enough to guarantee the above identifiability condition, as we show below. 

\section{Core Results}

\subsection{MCInfoNCE Learns the Correct Posterior}

The main result of this paper is a proof that the only minimizer of $\mathcal{L}_\text{MCInfoNCE}$ are probabilistic embeddings $Q(Z|X)$ that are equal to the true posteriors of the generative process, up to a general rotation of the whole embedding space $\mathcal{Z}$. This has some technical assumptions like that the probabilistic embedding distribution must be the same family as the true posterior, in this case vMF distributions, for otherwise it is impossible to exactly match it. We refer to the appendix for the full statement, conditions, and proof. This proof shows that the variances of probabilistic embeddings are not just training artifacts, but theoretically grounded.

\subsection{The Correctness is Robust to Violations of Assumptions}

We empirically verify this proof in a controlled experiment where a generator network with a randomly initialized posterior produces ambiguous data. We train a probabilistic embedding encoder using MCInfoNCE and find that its probabilistic embeddings are indeed equal to the generator posterior. This is robust to violations of the assumptions, like a different distribution family, a too low or too high dimensional embedding space, or even a generator without uncertainty, in which case MCInfoNCE correctly converges to Dirac probabilistic embeddings, that is, deterministic embeddings. We further test the vMF loss from the previous chapter and find that it also leads to correct posteriors, showing that several probabilistic embedding approaches learn correct uncertainties. However, this is no trivial property, as other losses like hedged instance embeddings (HIB, \citeauthor{oh2018modeling}, \citeyear{oh2018modeling}) do not provide the correct posteriors.

\subsection{The Learned Uncertainties are Aleatoric Uncertainties}

The theoretical formulation of the data-generating process hinted at the idea that these uncertainties are intrinsic to the image and that even the best model, the true posterior, could not reduce them. This is known as aleatoric uncertainty \citep{hullermeier2021aleatoric}. To investigate this experimentally, we apply MCInfoNCE to CIFAR-10H \citep{peterson2019human}, an image dataset with around 50 annotations per image. The entropy of these annotations serves as a proxy for the irreducible aleatoric uncertainty. We find that our probabilistic embeddings' uncertainties are indeed correlated to those human ones. Similarly, they are correlated to the amount of aleatoric uncertainty we synthetically induce in images by cropping them, thereby losing information. This is first evidence that we can learn the aleatoric uncertainty of images and their representations. 

\section{Discussion}
This work focussed on RQ2, understanding what our uncertainties about latents represent. To the best of my knowledge, it is the first paper to find that uncertainties in latent spaces are not just theoretical artifacts of variational training but have a real-world justification and show consistent behaviour. One detail that underlines this is that the uncertainties are trained from a randomly initialized MLP without any prior bias that could explain away its behaviour. This is an issue in the previous chapter (and the literature it followed \citep{scott2021mises,ko2021learning,ranjan2017l2}), where the representation vector norm we used to parametrize the uncertainties is nowadays suspected to be a mere fragment of cross-entropy training \citep{kang2023deep}. Beyond RQ2, this chapter also added to RQ1 by giving a new approach to learn uncertainties about latent representations, this time self-supervised, and opened ways for RQ3 by pioneering evaluations that test the uncertainties about unobservable latents against observable ground-truths. 

One limitation is that this work is limited to vMF posteriors. An extension to different exponential families as in \citet{dml_inversion} or even mixture densities would have been possible because the proof's arguments still hold: 1) MCInfoNCE is a cross-entropy that when optimized equalizes a certain expected positivity score between the generative process (and its posterior embeddings) and the predicted embeddings. This proof does not use the vMF assumption. And 2), this expected positivity can only be equal to the generative process's expected positivity if all embeddings match the true posteriors. The uniqueness argument of 2) would still hold for other posterior families, except that permutations of mixture components and other invariances in distribution parameters could add technical corner-cases. Formalizing these corner cases would have been very time-consuming without adding interesting proof techniques or novel understanding, so we decided to present only the core result on vMFs.

A second limitation is the scale of the experiments. Our first experiment was necessarily a toy experiment, because we needed full control over the data-generating process which is unknown in real-world data. The second experiment on real-world data, however, could have been on a larger scale. MCInfoNCE is scalable because it only adds a lightweight MLP head and 16 Monte-Carlo samples at a late model layer with near diminishing runtime and memory costs. The limiting factor here is that only CIFAR-10H (or datasets of similar scale \citep{schmarje2022one}) provide sensible ground-truths to compare our uncertainties against. At the time of publication, the literature lacked metrics and benchmarks to evaluate representation uncertainties, which hindered their development and scaling. This changed with the paper we present in the next chapter.

%% file: paper3.tex
\chapter{The URL Benchmark for Representation Uncertainties}
\label{chap:neurips}

\input{paper3/main}

%% file: paper3/main.tex
\textbf{Michael Kirchhof}, Bálint Mucsányi, Seong Joon Oh, and Enkelejda Kasneci. URL: A representation learning benchmark for transferable uncertainty estimates. \textit{Neural Information Processing Systems Track on Datasets and Benchmarks (NeurIPS D\&B)}, 2023.

\section{Prologue}

"That's neat, but does it scale?", inquired Enkelejda. She did not mean MCInfoNCE, for scaling it to a larger dataset and architecture would still fit on consumer-grade GPUs. Instead, the challenge in scaling the previous results was the evaluation protocol. What metric could we compare the learned uncertainties against if human uncertainty ground-truths are not available? I started forging and comparing several metrics. After weeks of tinkering, there finally was one metric to rule them all, one metric to find the best methods, one metric to scale them and in benchmarks develop them.\footnote{Inspired by \textit{The Lord of the Rings, The Fellowship of the Ring}, J.R.R. Tolkien, 1954, George Allen and Unwin.} This metric, the R-AUROC, would allow to benchmark arbitrary representation uncertainty methods on arbitrarily large datasets -- if only I could implement them by the rapidly approaching NeurIPS deadline. I turned to Bálint Mucsányi, a student visiting the lab and inter alia an excellent code engineer, and asked "Do you want to write a NeurIPS paper?".

% How could we even benchmark this? 

% Human annotations would be expensive and not scale themselves.

% So what could we do? 

% Then, found a metric from theory, and in practice it has all the properties we wanted

% "This section broadens the view, in that it moves away from specific methods, and towards formalizing the problem itself."

% 

\section{Motivation}

The previous chapters have demonstrated the promises of representation uncertainties. But further progress can only be enabled by a large-scale benchmark. The key to such a benchmark is a metric that quantifies the performance of representation uncertainties and is 1) scalable, 2) easy to implement, but 3) hard enough to be of longer term utility. Human annotations, as in the previous chapter, are not scalable as they have to be recollected for each dataset. Interventional metrics, like checking if uncertainties increase when an image is cropped or deteriorated, can be easily cheated by an overspecialized approach and are already saturated. In this section, we find a metric that fulfills all above criteria, and is even correlated with the human annotation gold-standard. We find it by broadening the view away from specific solutions and towards the problem that representation uncertainties address from a decision theory perspective.

\section{Methods}

We derive our metric, the R-AUROC, by reconsidering the problem from a decision theory perspective. In principle, uncertainties reflect the loss we expect when making a decision. In classification, when we give the decision "dog" with probability $80\%$, we quantify how high of a 0/1 correctness loss we expect. We can evaluate our uncertainty estimate of $80\%$ by comparing it to the actual 0/1 correctness on test data. In representation learning, our decision is the representation vector and a popular loss is the Recall@1. The Recall@1 measures if, when we embed all test samples, each representation's next neighbour is in the same class. This is also a 0/1 loss. So, when we give an uncertainty estimate about our decision, the representation, we evaluate whether it is predictive of this 0/1 correctness. To quantify this, we use the area under the ROC curve (AUROC) that tells if the uncertainties are predictive of the binary outcome variable. We name this the representation AUROC (R-AUROC). The R-AUROC allows evaluating a broad range of approaches, including ones that give a variance estimate $u(x) \in \mathbb{R}$ instead of a probability $u(x) \in [0,1]$. It can be evaluated on any classification dataset without new annotations, overcoming the previous hurdle, and can be added to existing representation learning benchmarks in four lines of code, thereby taking the practical hurdle for the field. 

The R-AUROC has another advantage inherited from representation learning: We do not need to know the classes at train time. They are added to the Recall@1, and hence the R-AUROC, at test time. This allows testing the representation correctness not just on seen but also on unseen datasets. We leverage this to test the transferability of uncertainty estimates on distribution shifts beyond previous benchmarks on robustness to corruptions \citep{galil2023a,ovadia2019can}. We train uncertainty estimators on ImageNet-1k \citep{deng2009imagenet} and evaluate them on three zero-shot datasets using the R-AUROC. This allow judging which approaches learn a notion of uncertainty that is transferable, paving the way for pretrained uncertainties.

The remaining details of the benchmark protocol are specified in the appendix. The core idea is to train eleven uncertainty estimators from the probabilistic embeddings from \cref{chap:eccv,chap:icml} to ensembles, determine their optimal hyperparameters via Bayesian optimization on a validation set, and test them via the zero-shot R-AUROC. To ensure a fair comparison, we reimplement all approaches as an extension of the \texttt{timm} \citep{rw2019timm} library. To move towards scalability, another catalyst for future pretrained uncertainties, we use both ResNet 50 \citep{resnet} and medium-sized Vision Transformers (ViT Medium, \citeauthor{vit}, \citeyear{vit}) as model backbones. Together, this comprises the uncertainty-aware representation learning (URL) benchmark.

\section{Core Results}

\subsection{The R-AUROC Metric Correlates with Gold Standard Metrics}

\begin{figure}[t]
    \centering \vspace{-4mm}
    \includegraphics[trim={0 0 6.78cm 0},clip,scale=1.1]{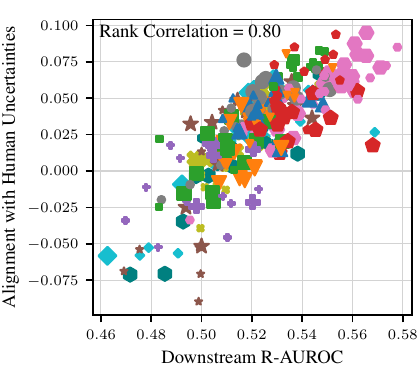} \hspace{1cm}
    \includegraphics[trim={7.2cm 0 0 0},clip,scale=1.1]{paper3/human.pdf}
    \caption{Dots represent all models we train with all approaches, hyperparameters, and backbones. Models with a higher R-AUROC reflect human uncertainties better (left) and behave better under uncertainty inducing transforms like cropping (right). This supports the R-AUROC empirically. Figure cited from the original paper \citep{kirchhof2023url}.}
    \label{fig:human}
\end{figure}

Before we start, we verify the integrity of our novel R-AUROC metric empirically by comparing it to existing uncertainty evaluation metrics. First, we use the gold standard from the previous section. That is, besides the R-AUROC, we track how well correlated the uncertainty estimates are with human annotator disagreements on five datasets \citep{schmarje2022one}, including the previous CIFAR-10H. \cref{fig:human} shows that these two metrics are highly correlated (rank correlation = 0.80) across all approaches, backbones, hyperparameters, and seeds we used in the benchmark. This means that whenever a model has a high R-AUROC, it also tends to score high on the gold standard human annotator metric (which is unavailable in most datasets). In the plot, even their random performance levels, 0.5 for the R-AUROC and 0 for rank correlation with human uncertainties, coincide. The same holds for an interventional metric that checks how often a smaller cropped version of an image receives a higher uncertainty estimate than its original (\cref{fig:human}, right). Last, the R-AUROC is also highly correlated with the widely used classification AUROC, when the latter is available on the seen classes of the ImageNet-1k validation set (see appendix). These experiments demonstrate that the R-AUROC judges uncertainty estimates consistent with previous gold standards, while being simpler to compute and available on arbitrary, even unseen, classification datasets.

\subsection{Approaches from Previous Chapters are Among the Best}

\begin{figure}[t]
    \centering \vspace{-1mm}
    \includegraphics[width=\textwidth]{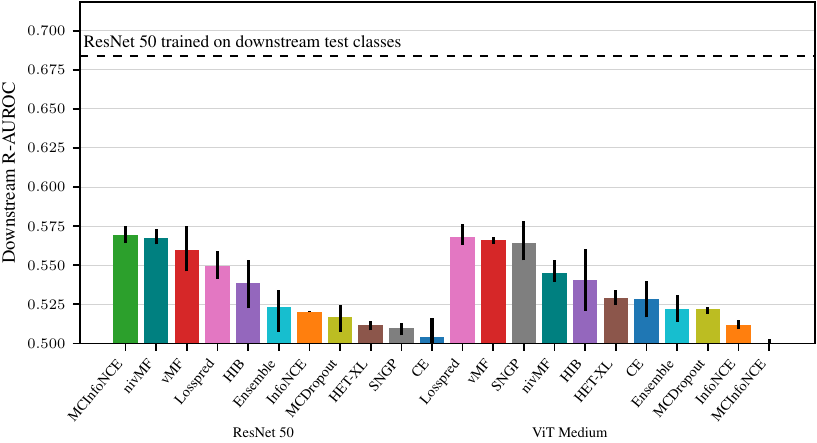}
    \caption{Methods from Chapters \ref{chap:eccv} and \ref{chap:icml}, MCInfoNCE, nivMF, and vMF, give among the best transferable uncertainties. Loss prediction also stands out, which we further investigate in \cref{chap:icml2}. Bars indicate minimum and maximum performance across three seeds. Figure adapted from the original paper \citep{kirchhof2023url}.}
    \label{fig:benchmark}
\end{figure}

We now use the R-AUROC to evaluate the methods from the previous chapters, MCInfoNCE, nivMF, and vMF, on a large scale and compare them to contemporary methods. \cref{fig:benchmark} shows that they are the best three approaches on ResNets and among the best on ViTs. A second approach, loss prediction, which we imported in this paper from regression literature \citep{upadhyay2023posterior,levi2022evaluating,laves2020well,Yoo_2019_CVPR}, also shows stable performance across both models. The plot additionally highlights that the R-AUROC is far from being saturated: As a reference, we trained a ResNet 50 with cross-entropy loss on the downstream datasets to obtain an upper bound on the performance. This is a loose bound since the zero-shot approaches do not know the precise downstream classes and can not use different uncertainty estimators on each downstream dataset, but it shows that there is room for improvements. In \cref{chap:icml2}, we reduce this gap by using URL to develop a new state-of-the-art.

\subsection{Uncertainty Quantification Sometimes Degrades the Main Task}

\cref{fig:benchmark} reveals another detail: MCInfoNCE is performant on ResNet 50, but not on ViT Medium. This is not a bug in implementation, but the result of a conflict of the intertwined representation learning and uncertainty estimation tasks. Although MCInfoNCE has one joint optimum, in practice the predicted (mean) representations and the predicted uncertainties drive the backbone into different gradient directions, and here the representatons' gradients were orders of magnitude stronger. This conflict is not exclusive to MCInfoNCE. In the main paper, we find that 15 of the 22 approaches have a trade-off when optimizing for Recall@1 versus for R-AUROC. We solve this in \cref{chap:icml2}.

\section{Discussion}

% Research questions
This paper answers RQ3 by providing a both theoretically funded and empirically well behaving metric to evaluate representation uncertainties. We developed it with RQ4 in mind, ensuring that it can be scaled to ImageNet and beyond. We also found that an approach originating from regression, loss prediction, achieves strong performance, adding to the methods sought after in RQ1. It takes a less variational, more direct approach at learning our desired representation uncertainties. This is why we further investigate unlocking its full potential in the next chapter, with a special emphasis on the gradient conflict we uncovered in this benchmark.

%% file: paper4.tex
\chapter{Pretrained Representation Uncertainties}
\label{chap:icml2}

\input{paper4/main}

%% file: paper4/main.tex
\textbf{Michael Kirchhof}, Mark Collier, Seong Joon Oh, and Enkelejda Kasneci. Pretrained visual
uncertainties. arXiv preprint arXiv:2402.16569, 2024. Under submission.

\section{Prologue}

"We need to talk!", I said, excitedly. I had just met Mark Collier at ICML. He had been working on the same problem of scalable uncertainties, had come to the same solution, probabilistic embeddings \citep{collier2023massively}, and, I figured, now faced the same hurdle. He had a look at my preliminary analysis of the gradient conflicts. A 15 minutes coffee break became a 1 hour lunch break became regular meetings with Enkelejda and Joon. We were determined to scale our previous efforts up, while cutting away any complexity that did not lead to measurable improvements on the URL benchmark. Ultimately, this chapter compiles the findings on all above research questions into one downloadable plug-and-play model. 

\section{Motivation}

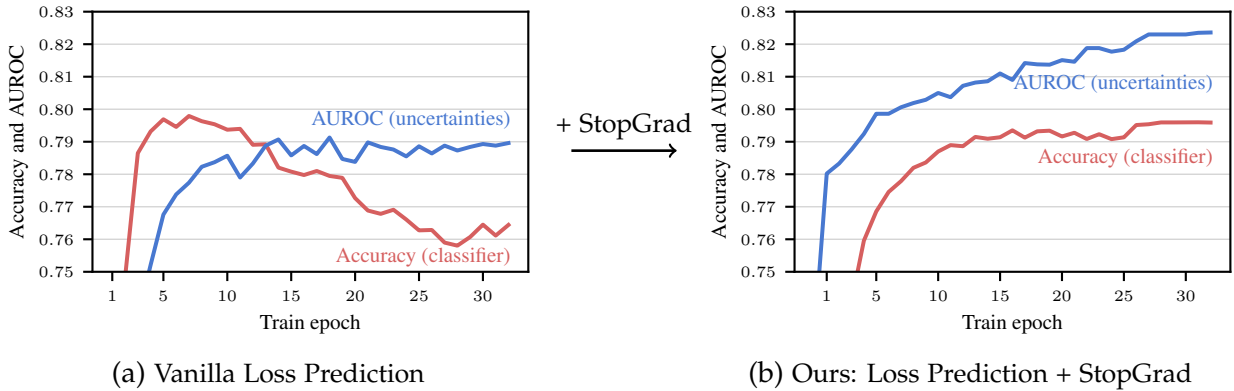
\begin{figure}[t]
\centering
\begin{subfigure}[b]{0.43\linewidth}
  \centering
  \includegraphics{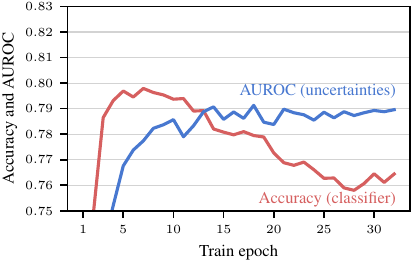}\vspace{0.5mm}
  \caption{\footnotesize Vanilla Loss Prediction}
  \label{fig:sub1}
\end{subfigure}\hfill% 
\begin{subfigure}[b]{0.43\linewidth}
  \centering
  \includegraphics{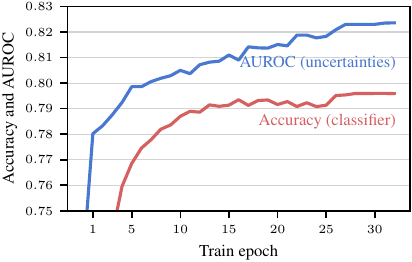}\vspace{0.5mm}
  \caption{\footnotesize Ours: Loss Prediction + StopGrad}
  \label{fig:sub2}
\end{subfigure}
  % Add arrow and text
  \begin{tikzpicture}[overlay, remember picture]
    \draw[->, line width=1pt] (-9.0,3.2) -- (-7.6,3.2);
    \node[above, text centered] at (-8.3,3.2) {\footnotesize + StopGrad};
  \end{tikzpicture} 
  
  \vspace{-1mm}
\caption{There is a conflict between the uncertainty and the classification objective when pretraining on ImageNet-1k, deteriorating both performances. A StopGrad resolves this conflict, enabling stable and scalable training. Figure cited from the original paper \citep{kirchhof2024pretrained}.}
\label{fig:test}
\end{figure}

The previous chapters have shown that our representation uncertainties are scalable and learn transferable notions of uncertainty. The last challenge is to scale them to large pretraining datasets, so that they can be deployed in a zero-shot plug-and-play manner by downstream practitioners. In particular, the pretrained model's representation uncertainties should (i) not interfere with the main pretraining or downstream task, (ii) generalize to zero-shot downstream datasets, (iii) flexibly adjust to any (downstream) task, (iv) have minimal compute overhead, and (v) converge stably to ensure scalability.

The main remaining hurdles are desiderata (i) and (v) because of the gradient conflict we discovered in \cref{chap:neurips}. As portrayed in \cref{fig:sub1} for loss prediction, where uncertainty estimation and representation learning are two distinct losses, the uncertainty objective hurts the performance of the pretrained model's main objective, transferable representations, and vice versa. This is because the gradients flowing back from both task heads attempt to change the backbone in interfering directions. This was so far avoided by stopping early, roughly around epoch 12 in the figure. However, early stopping prohibits training on large pretraining corpora. This chapter presents our solution to solve the conflict, scale up the training, and, finally, provide pretrained representation uncertainties for computer vision, as we set out to find in this thesis.

\section{Methods}

We decide to base our model on the loss prediction method from \cref{chap:neurips} over the equally well performing probabilistic embeddings, which we discuss further in the discussion below. Loss prediction stems from a decision theory perspective. The (in-)correctness of any task is defined by its loss function. Hence, to provide uncertainty estimates $u(x) \in \mathbb{R}$ that predict incorrectness, we predict the (gradient-detached) loss at each input \citep{Yoo_2019_CVPR}. As in the previous two chapters, this uncertainty estimation is realized by a lightweight MLP head added to a pretrained model. To a practitioner, this results in a simple dual-output API.
\begin{lstlisting}
embedding, uncertainty = pretrained_model(input)
\end{lstlisting}
A big hurdle is the gradient conflict. Although we have experimented with techniques like PCGrad \citep{yu2020gradient} to resolve it, the best performing and simplest solution is to place a \texttt{StopGrad} between the uncertainty MLP head and the model backbone. This strictly ensures the non-interference principle (i) and improves not only the main objective, but also the uncertainty performance, as shown in \cref{fig:sub2}.

The last challenge was to train on large pretraining corpora, here ImageNet-21k \citep{deng2009imagenet}, with large Vision Transformer backbones under limited compute. The solution is also enabled by \texttt{StopGrad}: Since \texttt{StopGrad} ensures that the backbone and classifier head are completely independent from influences of the uncertainty module, their training is orthogonal. We first load a pretrained checkpoint for the backbone (and classifier), and then cache the representations $e(x)$ throughout the whole training process once (all epochs including their random augmentations). Then we train the uncertainty head, which only needs to load the representations as inputs and the class labels as targets from disk. This increases the training throughput by a factor of 180x, enabling to train the uncertainty head of a ViT-Large for seven ImageNet-21k epochs (92 million samples) in 2:26 hours on a single V100 GPU, as opposed to 18 days when loading images $x$ from disk. 

We report more possible methodological enhancements in the main paper, but as negative results. None of them substantially increases the performance on the URL benchmark. We remove them to maintain the simplest possible approach.

\section{Core Results}

\subsection{Pretrained Visual Uncertainties Transfer Across Datasets}

\afterpage{%
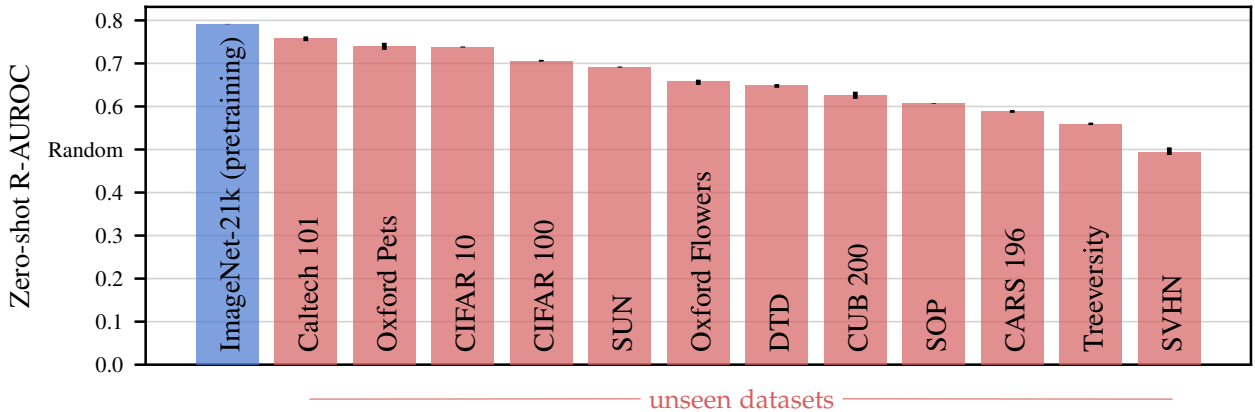
\begin{figure}[t]
    \centering
    \begin{tikzpicture}
    \node[anchor=south west,inner sep=0] (image) at (0,0) {\includegraphics[width=\textwidth]{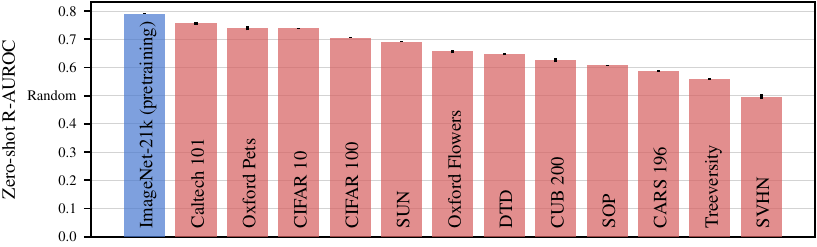}};
    \begin{scope}[x={(image.south east)},y={(image.north west)}]
        %\draw[help lines,xstep=.01,ystep=.01] (0,0) grid (1,1);
        %\draw[help lines,xstep=.1,ystep=0.1,color=black] (0,0) grid (1,1);
        \node[anchor=north,inner sep=1mm] (text) at (0.589,0) {\footnotesize \textcolor{red}{unseen datasets}};
        \draw[red] (text.west) -- (0.243, -0.048);
        \draw[red] (text.east) -- (0.934, -0.048);
    \end{scope}
    \end{tikzpicture}
    \caption{Pretrained uncertainties transfer to various downstream datasets, as measured by the R-AUROC. Bars indicate minimum and maximum performance across three seeds. Figure cited from the original paper \citep{kirchhof2024pretrained}.}
    \label{fig:generalize}
\end{figure}
}
The performance of our enhanced pretrained uncertainties exceeds that of the previous approaches on the URL benchmark, even when we use the smaller ImageNet-1k dataset for pretraining. In fact, the datasets in the URL benchmark (CUB 200 \cite{cub200}, SOP \citep{song2016deep}, and CARS 196 \citep{cars196}) are among the hardest due to their fine-grained and thus highly specialized classification task. \cref{fig:generalize} shows that our pretrained uncertainties generalize to other  natural image datasets, including those from the visual task adaptation benchmark \citep{zhai2020largescale}. This shows that our pretrained uncertainties behave as expected from a pretrained model, spanning the domain of the natural images pretraining dataset.

\subsection{Pretrained Uncertainties Represent Aleatoric Uncertainties}

When providing uncertainties, it is inevitable to specify which kind of uncertainties these are, commonly epistemic or aleatoric \citep{hullermeier2021aleatoric}. Epistemic denotes uncertainties about the correct choice of model parameters on unseen inputs, which can be reduced by collecting more similar inputs. Aleatoric are uncertainties in the data itself, e.g. a blurred or pixelated image, which are irreducible even with an expert or a Bayes-optimal model. We hypothesize that our pretrained visual uncertainties capture aleatoric uncertainty, without epistemic uncertainties. 

We find three pieces of evidence for this in the paper. First, ImageNet images where humans report ambiguity \citep{beyer2020we} receive a higher pretrained uncertainty estimate than images where they agree on a Dirac label, similar to \cref{chap:icml,chap:neurips}. Second, if we intervene on the images by cropping, but also by noising, blurring, or overlaying with grey boxes, uncertainties increase with the strength of intervention. Third, we find that uncertainty estimates on unseen datasets follow the same distribution as on the seen pretraining dataset, indicating the absense of epistemic influences.

These findings support our hypothesis that pretrained uncertainties model aleatoric uncertainty exclusively. This is a positive trait for a pretrained model, since it is intended to be deployed on unseen data where generally high epistemic uncertainty could drown out any aleatoric signal. It also provides one of the first methods that can disentangle the two uncertainties, which has been a recent effort in the field because it enables novel applications \citep{wimmer2023quantifying,mucsanyi2024benchmarking}.

\section{Discussion}

This work focussed on RQ4, overcoming remaining challenges to enable scaling, benefiting from the approaches and benchmarks we've built through the previous chapters and research questions. But it also adds new understanding about which uncertainties our representation uncertainties resemble, thereby contributing to RQ2. 

We made two major design choices in developing our pretrained visual uncertainties: Using loss prediction as a starting point, and applying \texttt{StopGrad}. Both have viable alternatives, which we discuss in the following two paragraphs.

We made our decision about the approach to base pretrained uncertainties from a problem-oriented perspective by introducing five desiderata meaningful to future practitioners. Both loss prediction and probabilistic embeddings have shown strong empirical performance in \cref{chap:neurips}, have theoretical foundations, and, with \texttt{StopGrad}, ensure non-interference, fulfilling desiderata (i), (ii), and (v). The biggest differences lie in the effort downstream users would have to make to adjust the pretrained model to their task of choice. In \cref{chap:icml}, we have seen that a simple blueprint can turn deterministic losses like InfoNCE into probabilistic ones like MCInfoNCE, maintaining their original properties and adding guarantees about the uncertainties. We are confident that this holds for further losses, but in comparison loss prediction is guaranteed by construction to adapt to any loss a downstream user may insert. As for compute, both methods output a scalar uncertainty $u(x) \in \mathbb{R}$ via an MLP head at inference time. However, at train time, MCInfoNCE requires Monte-Carlo samples whereas loss prediction uses the already computed loss value. This difference is small as sampling only happens in a late layer, but practitioners may be discouraged when remembering the large computational hurdle of sampling-based approaches like deep ensembles or MCDropout \citep{mucsanyi2024benchmarking}. Thus, we anticipate that pretrained uncertainties based on loss prediction will be more readily accepted outside the uncertainty quantification field.

The decision for \texttt{StopGrad} was based on Occam's razor. Instead of using \texttt{StopGrad} to ensure that the uncertainty training does not interfere with the backbone, we could have first trained the backbone and classifier head and then frozen it before training the uncertainty head. This stagewise training would have been functionally the same, but \texttt{StopGrad} unloads this implementation hurdle from practitioners. Further, it enables training both the main task head and the uncertainty head at the same time (as in \cref{fig:test}), providing a fail-safe mechanism for future users. Second, we could have implemented a gradient disentanglement approach like PCGrad to overcome the interferences. However, besides not working empirically in preliminary experiments, this would complicate training and add a dependency to be tuned. We encourage future researchers to reassess this point when the multi-task learning community finds new, robust algorithms.

%% file: conclusion.tex
\chapter{Discussion}
\label{chap:conclusion}

Uncertainties are often thought of as probabilities over output classes or intervals of the target variable in regression. These uncertainties are specific to each individual task. To provide uncertatinties that are more independent of the task, we attached uncertainties to representations. \cref{chap:eccv} demonstrated how to achieve this by simple adjustments to representation learning. \cref{chap:icml} showed that these uncertainties about latent representations indeed have a provable notion of correctness. \cref{chap:neurips} provided a benchmark to quantify how practically correct different methods for representation uncertainties are. \cref{chap:icml2} brought these findings to a large scale and developed a pretrained model whose representation uncertainties transfer across datasets. 

Besides this transferability, representation uncertainties also enable novel applications, which we outline below. Further, we comment on how our pretrained uncertainties are a starting point for specialized uncertainties, in which we see fruitful ends in future of uncertainty quantification research.

\section{Applications}

Representation uncertainties add a new dimension to representations that opens up novel applications. Having provided a downloadable model for representation uncertainties in \cref{chap:icml2}, we expect future research to explore the multitude of applications that representation uncertainties enable. We outline some applications below, some of which we already investigated in the main papers in the appendix, whereas others are given as inspiration for future researchers. 

\begin{figure}[t]
    \centering
    \includegraphics[scale=1.4]{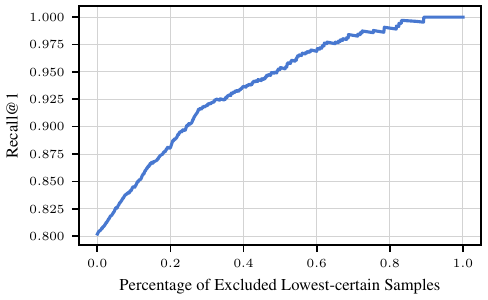} 
    \caption{When the model rejects inputs where it predicts a high uncertainty, it can achieve a high accuracy on the remainder of the data. This abstained prediction enables deploying models in situations with a high desired accuracy. Performance of MCInfoNCE trained on CIFAR-10. Figure cited from the original paper \citep{kirchhof2023probabilistic}.}
    \label{fig:abstained}
\end{figure}

We start with a traditional application of uncertainties, selective prediction \citep{el2010foundations,tran2022plex,galil2023what}. In selective prediction, the estimated uncertainty of each input is compared to a threshold and if the uncertainty is too high, we refuse to predict on this input. \cref{fig:abstained} shows that when increasing this threshold and rejecting more of the samples that the model considers uncertain, the accuracy on the remaining samples in fact increases. While machine learning may not be able to handle all inputs, this allows giving automated decisions at least for certain ones at a high accuracy. Remaining samples can, e.g., be asked to be re-taken or handed over to humans for inspection, as proposed by \citet{tran2022plex}.

The paradigm of selective prediction can also be applied to retrieval. In retrieval, the user inputs an image (or in multimodal settings, a text \citep{chun2021probabilistic,upadhyay2023probvlm}) and we output matching images from our database by comparing the representations of inputs and the database. Uncertainties can be added in two ways: We can reject inputs that are uncertain, and we can remove images with high representation uncertainties from our database to prevent matching them to any input. In the main paper of \cref{chap:icml2}, we show that these enhancements decrease the retrieval error, both on databases the model was trained on (10\% reduction each) and on databases where it gives zero-shot uncertainties (14\% and 17\% each). This marks low-hanging performance improvements and paves the way for safer retrieval.

\begin{figure}[t]
    \centering
    \begin{tikzpicture}
    \node[anchor=south west,inner sep=0] (image) at (0,0) {\includegraphics[trim=0cm 0.4cm 0cm -0.35cm, width=0.65\textwidth,page=2]{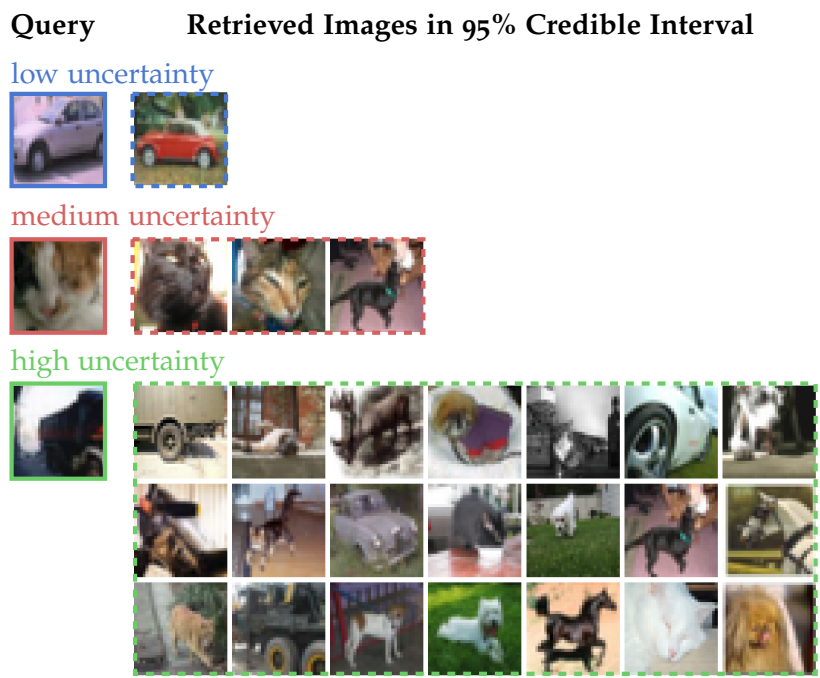}};
    \begin{scope}[x={(image.south east)},y={(image.north west)}]
        %\draw[help lines,xstep=.1,ystep=.1] (0,0) grid (1,1);
        %\foreach \x in {0,1,...,9} { \node [anchor=north] at (\x/10,0) {}; };
        \node [anchor=south west] at (-0.01,1.02) {\small \textbf{Query}};
        \node [anchor=south] at (0.57,1.02) {\small \textbf{Retrieved Images in 95\% Credible Interval}};
        \draw[blue,ultra thick] (0.005,0.951) rectangle (0.12,0.795);
        \draw[blue,ultra thick,dashed] (0.154,0.951) rectangle (0.269,0.795);
        \node [anchor=north west] at (-0.009,1.024) {\small \textcolor{blue}{low uncertainty}};
        \draw[red,ultra thick] (0.005,0.705) rectangle (0.12,0.547);
        \draw[red,ultra thick,dashed] (0.154,0.705) rectangle (0.512,0.547);
        \node [anchor=north west] at (-0.009,0.783) {\small \textcolor{red}{medium uncertainty}};
        \draw[green,ultra thick] (0.005,0.461) rectangle (0.12,0.3);
        \draw[green,ultra thick,dashed] (0.157,0.461) rectangle (0.995,-0.03);
        \node [anchor=north west] at (-0.009,0.538) {\small \textcolor{green}{high uncertainty}};
    \end{scope}
    \end{tikzpicture}
    \caption{When a user inputs an image whose representation is uncertain, we retrieve multiple images that may match the input. The size of the output set depends on the ambiguity of the input. Here, it is the 95\% highest density region of the input's probabilistic embedding, learned by MCInfoNCE on CIFAR-10. Figure adapted from the original paper \citep{kirchhof2023probabilistic}.}
    \label{fig:unc_retrieval}
\end{figure}

If one is hesitant to reject user queries, one can also react more softly to ambiguous retrieval inputs. \cref{fig:unc_retrieval} shows how we utilize the representation uncertainty, here probabilistic embeddings over the latent space from \cref{chap:icml}, to flexibly adjust how many possible matches we return to the user. If an input is clear, like the car in the top example, we return only a single sample, also a car. If the input is ambiguous, like in the lower examples, we return multiple matches, spanning images from several possible classes. This simple way to visualize the uncertainty is concurrently explored by \citet{upadhyay2023probvlm} and reminds of the current advances in conformal prediction \citep{angelopoulos2022gentle} where one outputs the set of all possible class labels to cover the true class with high probability. This similarity is no coincidence: Conformal prediction builds and calibrates these sets on the basis of score functions that indicate the uncertainty of every possible event. Our (pretrained) representation uncertainties are such score functions, enabling future advancements in zero-shot conformal prediction.

Another area that can benefit from our representation uncertainties is active learning. The most recent approaches \citep{mindermann2022prioritized,lahlou2023deup} seek samples that are not learned yet but of high quality. In other words, samples that have a high epistemic but low aleatoric uncertainty. To this end, they require estimators for aleatoric uncertainty that are not influenced by epistemic uncertainty. As we have seen in \cref{chap:icml2}, our pretrained representation uncertainties are among the first approaches to fulfill these criteria, simplifying active learning endeavours.

A similar strain of literature is dataset curation and handling noisy training signals \citep{pmlr-v202-ortiz-jimenez23a,marion2023less,sachdeva2024train,evans2024bad}. This challenge gained new interest with the current paradigm of using web-crawled, uncurated data to train large models \citep{schuhmann2021laion400m,tran2022plex}. Recent approaches find that removing low-quality data improves performance. Pretrained representation uncertainties capture precisely this, inputs with a generally low quality, and can be computed on the spot even for new data, enabling future use as dataset curators.

Last, representation uncertainties can be used in any approach that uses representations to visualize datasets, such as clustering \citep{JMLR:v9:vandermaaten08a,McInnes2018}. We have already seen in \cref{fig:clustering} how uncertainties enhance these plots to communicate uncertainties to practitioners, allowing to understand and debug datasets for more trustworthy machine learning.

\section{Specialized Uncertainties}

Throughout the thesis, the reader may have noted an increasing abstraction of our uncertainties. Whereas previous approaches commonly define uncertainties as, e.g., classification probabilities, \cref{chap:eccv,chap:icml} first abstract uncertainties away from the classification task and towards uncertainties about representations in general, independent of the specific task. \cref{chap:neurips} generalizes this further by going from probabilistic embeddings towards any uncertainty estimator that provides uncertainties about representations. Last, we settle on a loss-based interpretation of uncertainties. If uncertainties aim to estimate how wrong we are then the loss quantifies what wrongness precisely means in the task the practitioner is handling. This is the paradigm that our pretrained uncertainties use in \cref{chap:icml2}, with the intention that the pretrained uncertainties will adjust to the practitioner's loss once they are finetuned on downstream data.

This shows our main vision: Specializing uncertainties to individual tasks \citep{franchi2022muad,mucsanyi2024benchmarking}. This is a pragmatic generalization of the recent efforts in uncertainty disentanglement. Here, the field is currently moving from one-fits-all predictive uncertainty values \citep{gal2016dropout,lakshminarayanan2017simple} to disentangled aleatoric and epistemic uncertainties \citep{hullermeier2021aleatoric,valdenegro2022deeper,wimmer2023quantifying,mucsanyi2024benchmarking}. One unsolved issue in this framework is that epistemic uncertainty remains only vaguely defined \citep{der2009aleatory,jurgens2024epistemic}. We expect that a loss-based view will move uncertainty estimation forward by making it more explicit and more specialized to the tasks practitioners intend to solve with it. 

As examples, aleatoric uncertainty in classification becomes the remaining cross-entropy loss of the trained classifier. Epistemic uncertainty for outlier detection becomes a 0/1 loss of a binary OOD classification task. Density estimation is predicting a log likelihood loss. If there are further tasks a practitioner wants to use uncertainties for, they do not have to be fitted into the epistemic-aleatoric dichotomy, but can be defined as a precise task to be optimized by the uncertainty module. This makes uncertainty estimation more pragmatic and more explicitly optimizable since loss prediction is, in essence, just another regression task. We anticipate that this specialization enabled by abstraction will both simplify and unify future works on uncertainty estimation.

\section{Conclusion}

This thesis brought uncertainties in computer vision to the layer of representations. This has the advantage that they can be pretrained on a large scale and then transferred to new datasets and tasks. Besides these practical advances, we also explored the theoretical foundation of uncertainties about latents and how to benchmark them. We compiled all these theoretical and practical findings into one downloadable model in order to facilitate uncertainty quantification for researchers inside and, importantly, outside the field. This demonstrates our vision for the future of uncertainty quantification: We encourage researchers from inside the field to shape their sophisticated methods and findings into pragmatic answers to the pragmatic questions practitioners outside the field face. We expect that this will enable a widespread application of uncertainties, making trustworthy machine learning the norm.

%% file: appendix_arxiv.tex
\begin{appendix}
\chapter{A non-isotropic probabilistic take on proxy-based deep metric learning}
This appendix includes the full paper and appendix discussed in \cref{chap:eccv}, but is left empty in the arXiv version of this work for license reasons. The paper and appendix can be found standalone (recommended) or in the official submission of this thesis below.

Michael Kirchhof, Karsten Roth, Zeynep Akata, and Enkelejda Kasneci. A non-isotropic probabilistic take on proxy-based deep metric learning, 2022. First published in: Avidan, S., Brostow, G., Cissé, M., Farinella, G.M., Hassner, T. (eds) Computer Vision – ECCV 2022. ECCV 2022. Lecture Notes in Computer Science, vol 13686, pp 435–454, by Springer Nature. DOI: \url{https://doi.org/10.1007/978-3-031-19809-0_25}.

Michael Kirchhof. Uncertainties of Latent Representations in Computer Vision, 2024. PhD thesis, Universität Tübingen. DOI: \url{http://dx.doi.org/10.15496/publikation-98103}.

\chapter{Probabilistic contrastive learning recovers the correct aleatoric uncertainty of ambiguous inputs}
This appendix includes the full paper and appendix discussed in \cref{chap:icml}, but is left empty in the arXiv version of this work for license reasons. The paper and appendix can be found standalone (recommended) or in the official submission of this thesis below.

Michael Kirchhof, Enkelejda Kasneci, and Seong Joon Oh. Probabilistic contrastive learning recovers the correct aleatoric uncertainty of ambiguous inputs, 2023. First published in: \textit{Proceedings of the 40th International Conference on Machine Learning}, in \textit{Proceedings of Machine Learning Research} 202:17085-17104. Available from \url{https://proceedings.mlr.press/v202/kirchhof23a.html}. 

Michael Kirchhof. Uncertainties of Latent Representations in Computer Vision, 2024. PhD thesis, Universität Tübingen. DOI: \url{http://dx.doi.org/10.15496/publikation-98103}.

\chapter{URL: A Representation Learning Benchmark for Transferable Uncertainty Estimates}
This appendix includes the full paper and appendix discussed in \cref{chap:neurips}, but is left empty in the arXiv version of this work for license reasons. The paper and appendix can be found standalone (recommended) or in the official submission of this thesis below.

Michael Kirchhof, Bálint Mucsányi, Seong Joon Oh, and Enkelejda Kasneci. URL: A representation learning benchmark for transferable uncertainty estimates, 2023. First published in: \textit{Neural Information Processing Systems Track on Datasets and Benchmarks (NeurIPS D\&B)}, in \textit{Advances in Neural Information Processing Systems 36 pre-proceedings (NeurIPS 2023)}. Available from \url{https://proceedings.neurips.cc/paper_files/paper/2023/hash/2d421cd0e763f9f01958a30bace955bf-Abstract-Datasets_and_Benchmarks.html}. 

Michael Kirchhof. Uncertainties of Latent Representations in Computer Vision, 2024. PhD thesis, Universität Tübingen. DOI: \url{http://dx.doi.org/10.15496/publikation-98103}.

\chapter{Pretrained Visual Uncertainties}
This appendix includes the full paper and appendix discussed in \cref{chap:icml2}, but is left empty in the arXiv version of this work for license reasons. The paper and appendix can be found standalone (recommended) or in the official submission of this thesis below.

Michael Kirchhof, Mark Collier, Seong Joon Oh, and Enkelejda Kasneci. Pretrained visual
uncertainties, 2024. First published in Arxiv. Available from https://arxiv.org/abs/2402.16569.

Michael Kirchhof. Uncertainties of Latent Representations in Computer Vision, 2024. PhD thesis, Universität Tübingen. DOI: \url{http://dx.doi.org/10.15496/publikation-98103}.

\end{appendix}